\documentclass[letterpaper, 10 pt, conference]{ieeeconf}  % Comment this line out if you need a4paper

\IEEEoverridecommandlockouts                              % This command is only needed if 
\usepackage{amsmath} % assumes amsmath package installed
\usepackage{amssymb}  % assumes amsmath package installed

\usepackage{graphicx}
\usepackage[hang]{subfigure}
\usepackage{booktabs}
\usepackage{scalerel}

\newcommand{\cS}{\mathcal{S}}
\newcommand{\cA}{\mathcal{A}}

\newcommand{\R}{\mathbb{R}}

\newcommand{\D}{\mathcal{D}}
\newcommand{\I}{\mathcal{I}}

\title{\LARGE \bf
Temporal Abstraction in Reinforcement Learning with Offline Data 
}

\author{Ranga Shaarad Ayyagari, Anurita Ghosh and Ambedkar Dukkipati%
\thanks{Email: \{rangaa, anuritaghosh, ambedkar\}@iisc.ac.in}%
\thanks{Department of Computer Science and Automation, Indian Institute of Science, Bangalore, India - 560012.}%
}

\begin{document}

\maketitle
\thispagestyle{empty}
\pagestyle{empty}

%%%%%%%%%%%%%%%%%%%%%%%%%%%%%%%%%%%%%%%%%%%%%%%%%%%%%%%%%%%%%%%%%%%%%%%%%%%%%%%%
\begin{abstract}
 Standard reinforcement learning algorithms with a single policy perform poorly on tasks in complex environments involving sparse rewards, diverse behaviors, or long-term planning. This led to the study of algorithms that incorporate temporal abstraction by training a hierarchy of policies that plan over different time scales. The options framework has been introduced to implement such temporal abstraction by learning low-level options that act as extended actions controlled by a high-level policy. The main challenge in applying these algorithms to real-world problems is that they suffer from high sample complexity to train multiple levels of the hierarchy, which is impossible in online settings. Motivated by this, in this paper, we propose an offline hierarchical RL method that can learn options from existing offline datasets collected by other unknown agents. This is a very challenging problem due to the distribution mismatch between the learned options and the policies responsible for the offline dataset and to our knowledge, this is the first work in this direction. In this work, we propose a framework by which an online hierarchical reinforcement learning algorithm can be trained on an offline dataset of transitions collected by an unknown behavior policy. We validate our method on Gym MuJoCo locomotion environments and robotic gripper block-stacking tasks in the standard as well as transfer and goal-conditioned settings.

\end{abstract}

%%%%%%%%%%%%%%%%%%%%%%%%%%%%%%%%%%%%%%%%%%%%%%%%%%%%%%%%%%%%%%%%%%%%%%%%%%%%%%%%
\section{INTRODUCTION}
\label{sec:intro}
In recent years, reinforcement learning has been used to solve a wide variety of sequential decision-making tasks. However, in many cases, policies learned using standard reinforcement learning algorithms struggle to plan over long-horizon tasks that consist of multiple steps. This is due to various reasons, such as a sparse reward signal that requires extensive exploration and the need to learn and compose diverse behaviors to execute complicated tasks.
Various techniques have been studied to tackle this problem of long-term planning over complex sub-tasks. One such technique is the Options framework, in which the agent learns multiple low-level policies, called options, controlled by a single high-level policy. The high-level policy chooses one of the options to be executed based on the current state, and the control is passed to the chosen option. This option then continues to be in control until it decides to terminate. The control is then passed to the high-level policy, which then chooses an option again, and so on.

Another important aspect of reinforcement learning algorithms in complex settings such as robotics is the issue of high sample complexity. This is especially relevant for hierarchical algorithms that need a lot of exploration to learn multiple levels of policies that interact with each other.

This led to the study of reinforcement learning algorithms that do not need access to an environment to explore and try different strategies. Instead, they just need access to a fixed dataset of samples collected from the environment by some possibly unknown behavior policy. Such methods come under the category of Offline Reinforcement Learning algorithms since they only use offline data and have no need for online access to the environment.

The offline setting poses new problems, such as distributional shift, lack of coverage of the environment, sub-optimality of the behavior policy, etc. While many algorithms that deal with these problems have been studied, the study of hierarchical algorithms that can deal with offline data is limited.

This work proposes a general framework to learn hierarchical reinforcement learning agents using only offline data. This can be used to convert an online hierarchical learning algorithm to work in the offline setting. We experimentally validate this framework in two environment settings and show that it can be used to extend online algorithms to operate only on offline data to learn hierarchical policies in their respective settings.

\begin{figure}[t]
\centering
\vspace{3mm}
\subfigure[Goal (i)]{\includegraphics[width=0.30\linewidth]{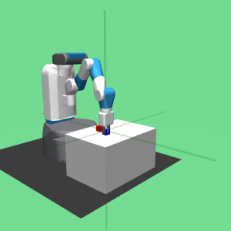}}
\subfigure[Goal (ii)]{\includegraphics[width=0.30\linewidth]{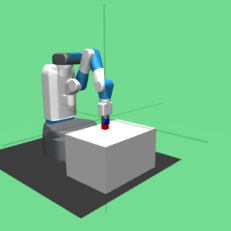}}
\subfigure[Goal (iii)]{\includegraphics[width=0.30\linewidth]{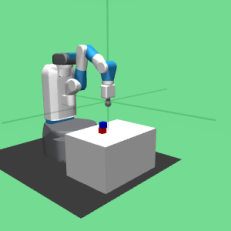}}
\caption{The three high-level goals in the robotics environment grasping the blue block, placing it on the red block, and returning to the end position, respectively. These actions have to be taken sequentially, necessitating a hierarchical agent with a high-level planner.}
\label{fig:goals_final}
\end{figure}

\subsection*{Contributions}

\begin{enumerate}
	\item We propose a general framework that can be used to convert an online hierarchical reinforcement learning algorithm for learning options to learn solely from offline data.
	\item We provide experimental validation of this framework on MuJoCo locomotion tasks in the standard and in the transfer setting.
	\item We then validate the suitability of our framework for planning tasks by learning a hierarchical agent in robot gripper block-stacking tasks.
	\item We provide ablations and analysis to show the relevance of each component of the framework to learn hierarchical agents that can leverage offline data for task transfer and high-level planning.
\end{enumerate}

\section{PRELIMINARIES}
\label{sec:preliminaries}

\subsection{Options Framework}
Consider a Markov Decision Process (MDP) defined by the tuple $(\cS, \cA, r, p, \rho, \gamma)$, where $\cS$ is the state space, $\cA$ is the action space, $r: \cS \times \cA \rightarrow \R$ is the reward function, $p$ is the probability transition function, $\rho$ is the initial state distribution and $\gamma$ is the discount factor.

The Options framework\cite{Sutton1999BetweenMDPsAndSemiMDPs} incorporates temporal abstraction into reinforcement learning agents by learning a hierarchy of policies that operate at different time scales. Low-level policies are learned that interact with the environment for a variable amount of time till termination, and a high-level policy chooses the low-level policy to which to give control.

Formally, the agent consists of a set $\Omega$ of options. Each option $\omega \in \Omega$ is defined by a tuple $(\I_\omega, \pi_\omega, \beta_\omega)$, where $\I_\omega$ is the set of initial states from which $\omega$ can be given control, $\pi_\omega$ is the actual policy corresponding to $\omega$, and $\beta_\omega$ is the termination function that decides when the option $\omega$ relinquishes control to the high-level policy. In most approaches, the initiation set $\I$ is the entire state space.

Extending the standard actor-critic approach, options framework approaches consider option-value functions $Q_\Omega(s, \omega), Q_U(s, \omega, a)$ and $V_\Omega(s)$, and the option-value function upon arrival $U(s', \omega)$ that incorporates the probability of the previous option $\omega$ terminating at state $s'$ upon arrival.

In our work, we develop offline algorithms on top of two option learning approaches. The first is the Multi-updates Option Critic (MOC) algorithm~\cite{Klissarov2021FlexibleOptionLearning}, which is an extension of the original Option-Critic algorithm~\cite{Bacon2017OptionCriticArchitecture}. The second is the Universal Option Framework (UOF) \cite{Yang2021HierarchicalRLWithUniversalPolicies}, which learns goal-conditioned low-level and high-level policies.

\subsection{Offline Reinforcement Learning}

The approaches discussed in the previous section require the agent to repeatedly interact with the environment to obtain diverse transitions. However, one might wish to avoid collecting any transitions and instead use an existing dataset of transitions collected by an unknown agent. This setting is called Offline Reinforcement Learning.

In this setting, the agent is given a dataset $\D$ that is a collection of transitions of the form
\[
\D = \left\{ (s_i, a_i, r_i, s_i', d_i) \right\}_{i=1}^N,
\]
where $s_i \in \cS$ is some state of the environment, $a_i$ is the action taken, $r_i$ is the reward obtained, $s'_i$ is the next state the environment transitioned into, and $d_i$ indicates whether this transition resulted in the termination of the episode.

Such a dataset could have been created by exploring the environment using an unknown behavior policy $\pi_b$ or some mixture of unknown policies. There are many issues with learning optimal policies from such an offline dataset. Evaluating a policy using transitions collected by a different policy leads to a distributional shift. Learning an optimal policy is prevented by limited coverage of the dataset due to the behavior policy being sub-optimal or narrow.

Many approaches have been proposed to deal with these issues, most of which incorporate pessimism into the learning process in various ways. In our work, we consider Model-Based Offline Reinforcement Learning (MOReL) \cite{Kidambi2020Morel}, in which an approximate pessimistic MDP model is learned from the dataset, and a model-based Natural Policy Gradient (NPG) algorithm is then used to learn an optimal policy in this model. The performance of such a policy is an approximate lower bound on its performance in the unknown real MDP.

\section{PROPOSED FRAMEWORK}
\label{sec:algorithm}

\begin{figure}%[t]
	\centering
    \includegraphics[width=\linewidth]{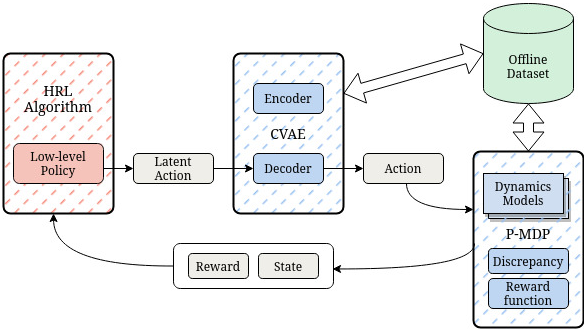}
	\caption{Meta-algorithm for learning an offline version of a hierarchical online algorithm. The CVAE and the environment models are learned from the offline dataset. These are then used to train the HRL algorithm, which operates in a pessimistic approximation of the actual environment and the latent action space of the CVAE.}
	\label{fig:alg}
\end{figure}

The offline dataset can directly be used to learn a standard reinforcement learning agent, sometimes even naively, because such an agent learns its own policy $\pi$ whose behavior can be inferred from the response of the environment to $\pi_b$.

However, when a hierarchy of policies needs to be learned, such as in the Option-Critic architecture, the performance of a high-level policy depends on the current set of options, and it is not possible to directly learn this value since the options keep changing and the dataset does not have information about how any current option would perform long-term. This is as opposed to the fact that the dataset contains information on rewards obtained by actions taken by a flat policy.

So to convert an arbitrary online hierarchical reinforcement learning algorithm into an offline algorithm, there is a compelling reason to learn the MDP model, more so than in the case of learning a flat agent. For this reason, we learn a P-MDP, a pessimistic approximation of the MDP that terminates with a reward penalty in regions of the state space where it is uncertain about the true MDP.

The pessimism of the P-MDP explicitly restricts the agent from exploring regions of the environment's state-action space that are not represented in the offline dataset and are hence unknown. Due to the penalty term, the agent is incentivized to plan ahead and stay within the confines of the offline dataset.

Another approach for restricting the agent from taking out-of-distribution actions is to explicitly learn the state-action distribution and only allow the agent to sample actions in support of this distribution. We also incorporate this approach by learning a Conditional Variational Auto-Encoder (CVAE) to model the latent space of actions conditioned on the state and allowing the low-level policy to only pick an action from this latent space.

\subsection{Standard setting}
\label{subsec:alg:standard}

Given an offline dataset $\D$ as described above, the first step is to construct a pessimistic MDP (P-MDP) that is a pessimistic approximate model of the real MDP. This is done by learning an ensemble $\{ T_k, R_k \}_{k=1}^K$ of dynamics models and reward functions. Each model $T_k$ takes as input a state-action pair and predicts the change in the environment state due to the action, while $R_k$ predicts the reward. The termination condition of the MDP is assumed to be known.

The $T_k$ are trained on $\D$ until convergence, and a discrepancy measure $M: \cS \times \cA \rightarrow \R$ is considered as a proxy to the epistemic uncertainty of the learned dynamics models. It takes a state-action pair as input and outputs the disagreement regarding the resultant next state among the dynamics models in the ensemble. For our experiments, we use the variance of the predictions as the discrepancy measure.

The P-MDP model is defined as follows. For sampling an episode from the P-MDP, one dynamics model $T_k$ and reward model $R_k$ are randomly chosen from among the ensemble. At each step, for current state $s$ and a given action $a$, the next state is calculated as $s' = s + T_k(\bar{s}, a)$ and the discrepancy is calculated as $M(s, a) = \textsf{Variance} \left( \left\{ T_j(\bar{s}, a) \right\}_{j=1}^K \right)$, where $\bar{s}$ is the normalized state. The reward is calculated as $R_k(\bar{s}, a)$.

If $M(s, a)$ is above a certain threshold, the episode is terminated, and a penalty is given to the agent. The threshold and penalty are hyperparameters that depend on the environment under consideration. This termination and penalty induce pessimism and deter the agent from wandering far from the support of the given offline dataset distribution.

The second step is to train a Conditional Variational Auto-Encoder (CVAE), consisting of an encoder $E$ and a decoder $D$. Each state-action pair $(s, a)$ in $\D$ is passed through the encoder to obtain the encoding $\bar{a} = E(\bar{s}, a)$. For a given state, an action can be sampled in $\cA$ by sampling an action $\bar{a} \sim \mathcal{N}(0, I)$ in the latent space and passing it through the decoder, giving $a = D(\bar{s}, \bar{a})$.

The CVAE is trained in the standard way, with the loss to be minimized being the sum of the reconstruction error and the KL divergence between the latent actions and the standard normal distribution.

Now, an arbitrary hierarchical RL Algorithm $H$ can be trained using a combination of the CVAE and the P-MDP as follows. When at state $s$, the low-level policy takes an action $\bar{a}$ in the state-conditioned latent action space of the CVAE. This is passed through the decoder to obtain the real action $a = D(s, \bar{a})$, which is then passed to the learned P-MDP. The P-MDP returns the next state and reward to $H$, which uses the transition as it sees fit.

In essence, the Algorithm $H$ operates in the approximate pessimistic MDP constructed from the offline dataset and is planned in the latent action space of the CVAE. It treats this setting as if it is an online environment and plans and learns as in the online setting without any modifications. The overall structure is given in Fig.~\ref{fig:alg}.

\subsection{Transfer setting}

An advantage of learning reusable low-level skills in a hierarchical agent is the transfer of such skills to solve a task different from the one the agent was trained on. If the low-level skills are reusable, then retraining the hierarchical agent on the new task will be more sample-efficient than retraining a flat agent.

In our work, we consider the setting where an offline dataset is given for a specific task in an environment, but the agent has online access to the environment and needs to learn a different task. We first train the hierarchical agent offline as described previously and then fine-tune it online on the new task.

The actions to be taken to solve the new task need not be the same actions that solve the original task. Therefore, while the agent learns to plan in the latent action space during the offline training, for online fine-tuning, after that, the decoder is discarded, and the agent is made to plan using the actual actions in the environment.

\begin{figure*}[h]
\centering
\vspace{2mm}
\subfigure[Hopper Medium]{\includegraphics[width=0.23\textwidth]{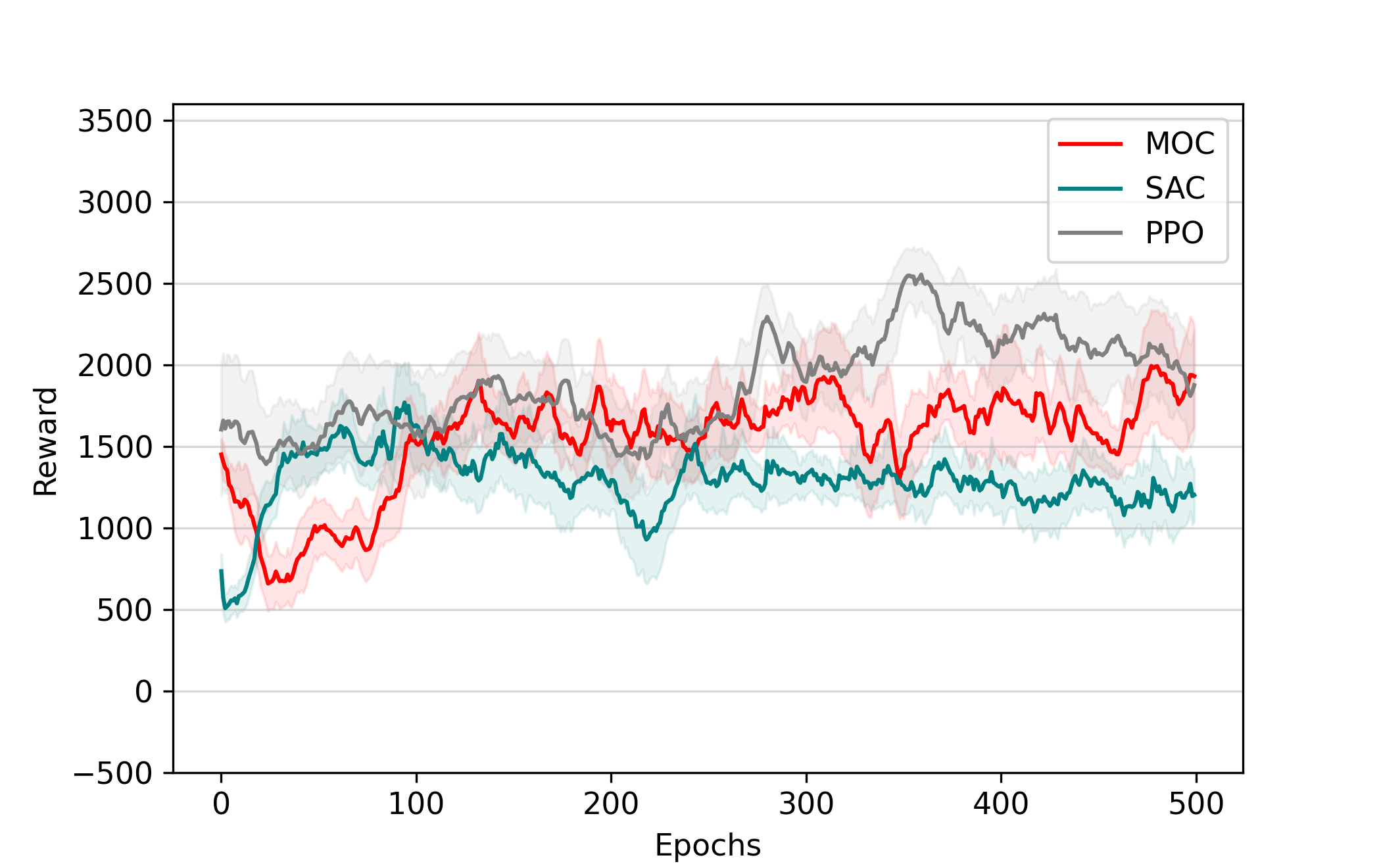}}
\subfigure[HalfCheetah Medium]{\includegraphics[width=0.23\textwidth]{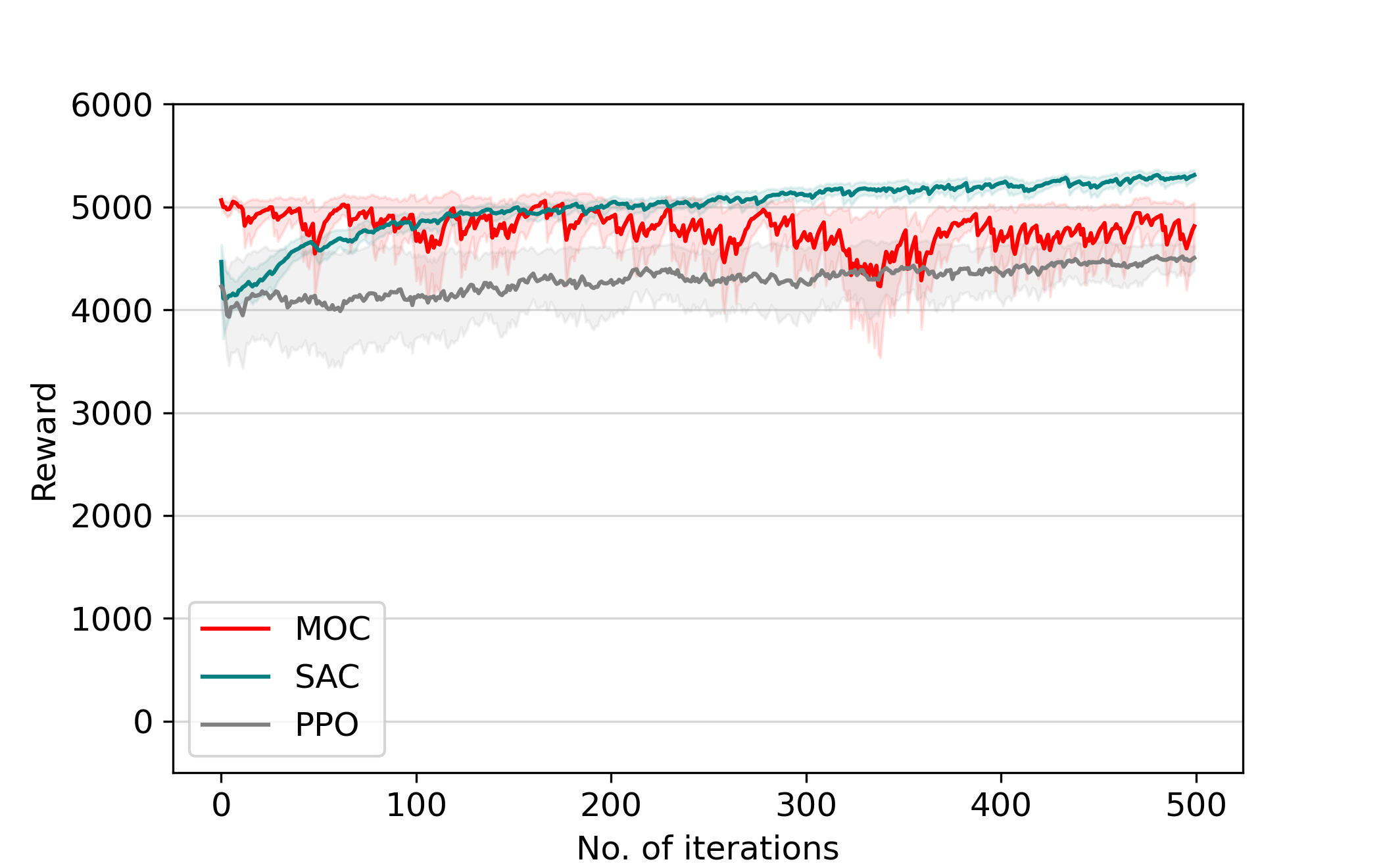}}
\subfigure[Hopper Medium-Expert]{\vstretch{0.8}{\includegraphics[width=0.23\textwidth]{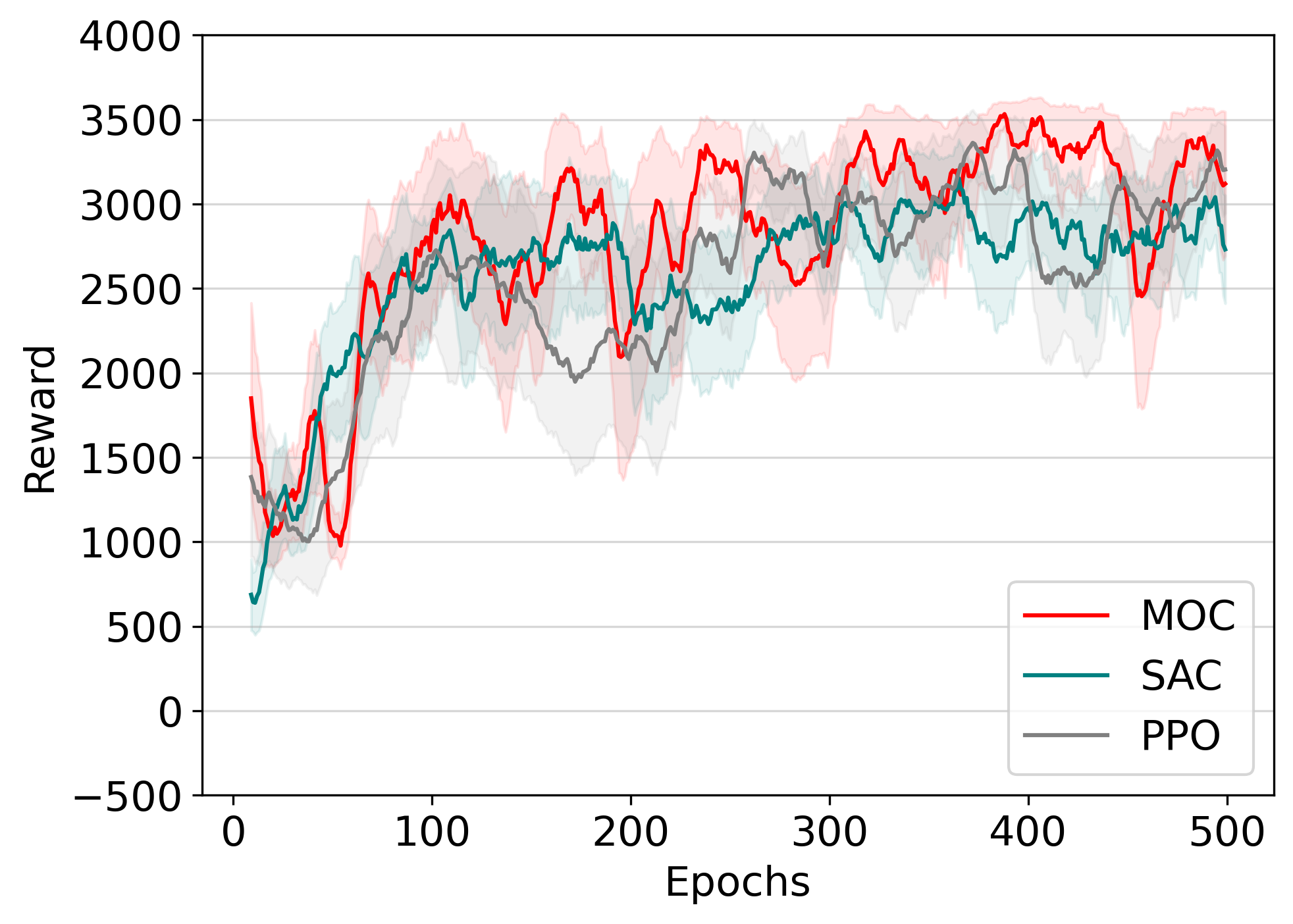}}}
\subfigure[HalfCheetah Medium-Expert]{\vstretch{0.8}{\includegraphics[width=0.23\textwidth]{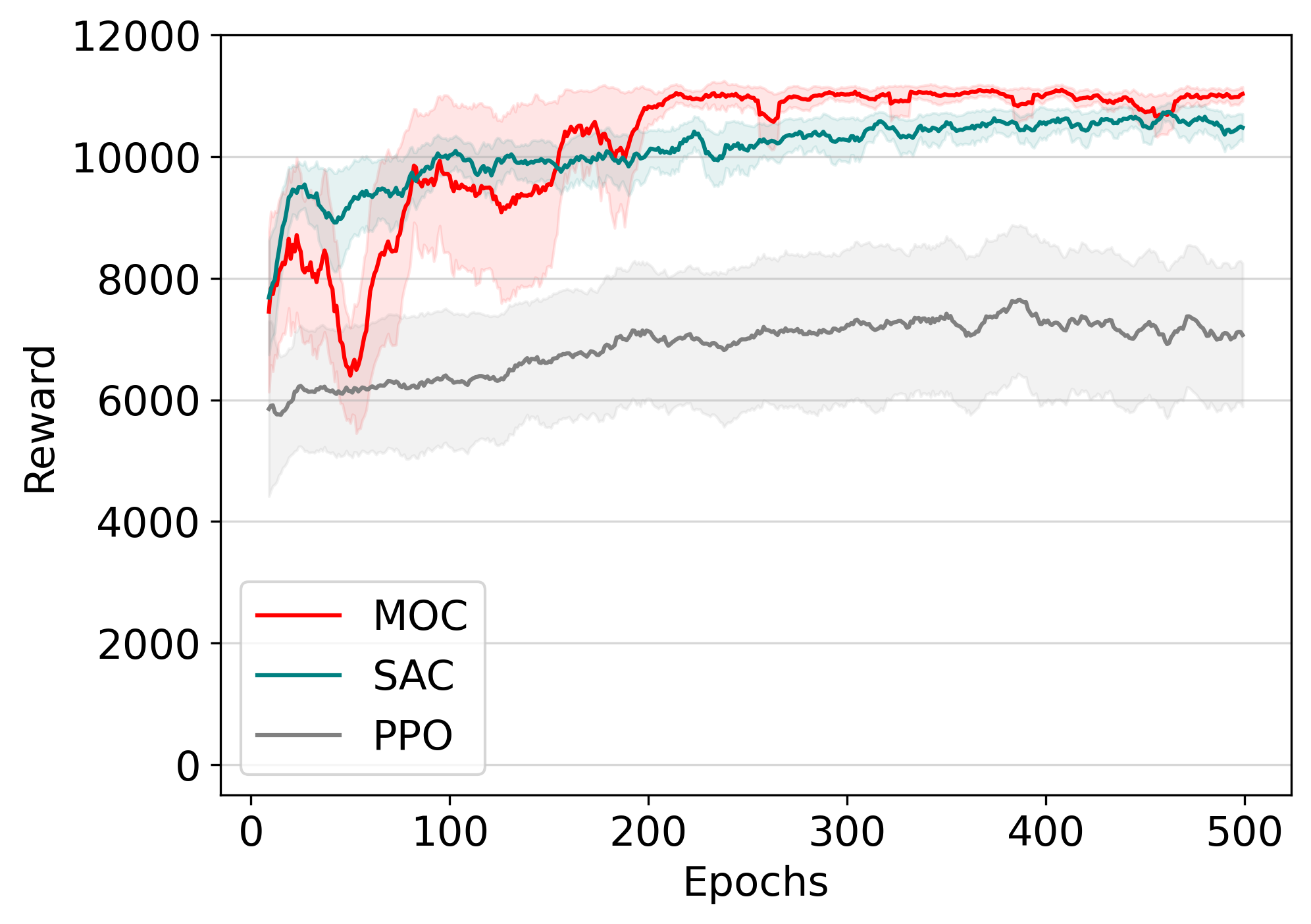}}} \\
\subfigure[Hopper Medium]{\includegraphics[width=0.23\textwidth]{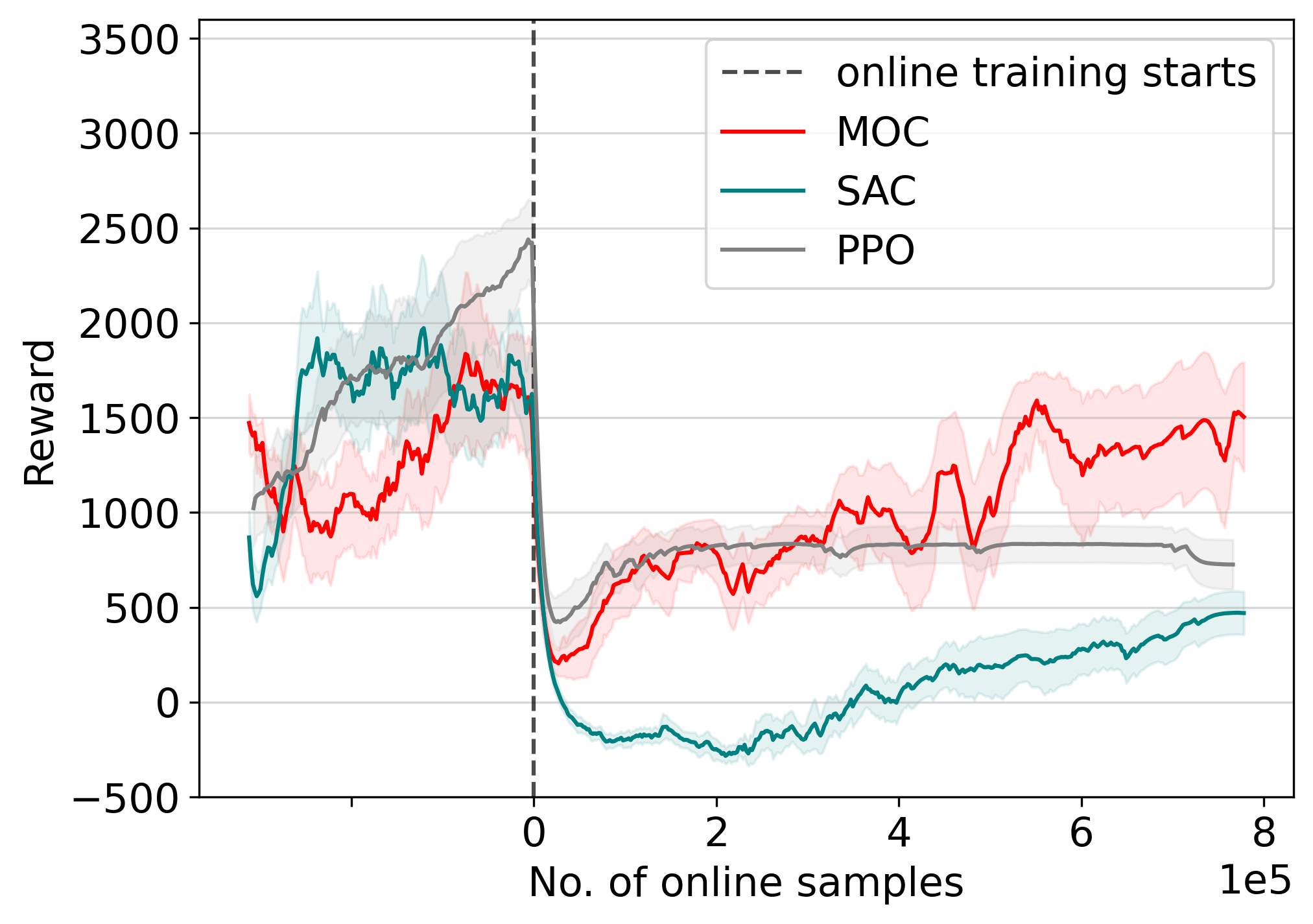}}
\subfigure[HalfCheetah Medium]{\includegraphics[width=0.23\textwidth]{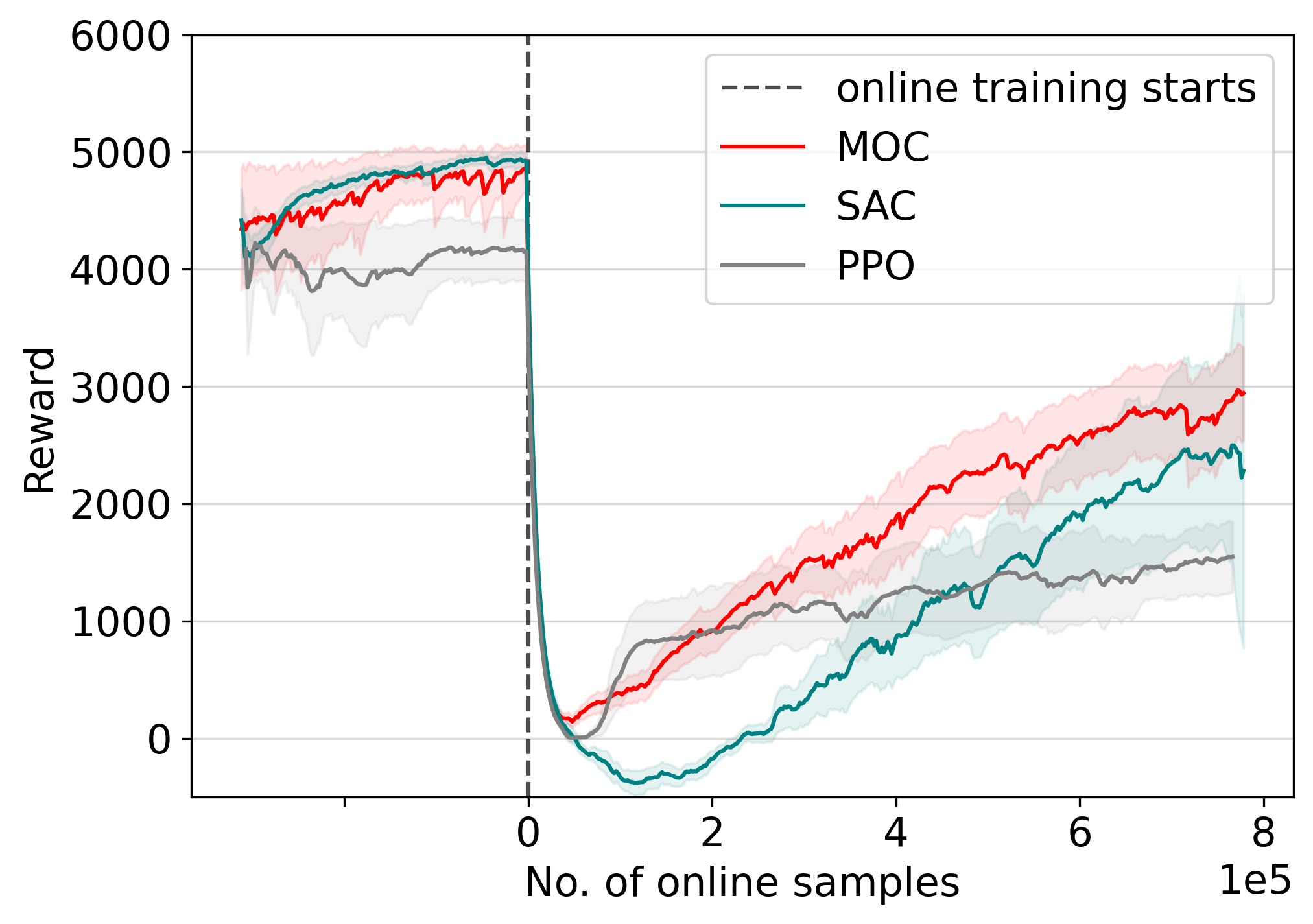}}
\subfigure[Hopper Medium-Expert]{\includegraphics[width=0.23\textwidth]{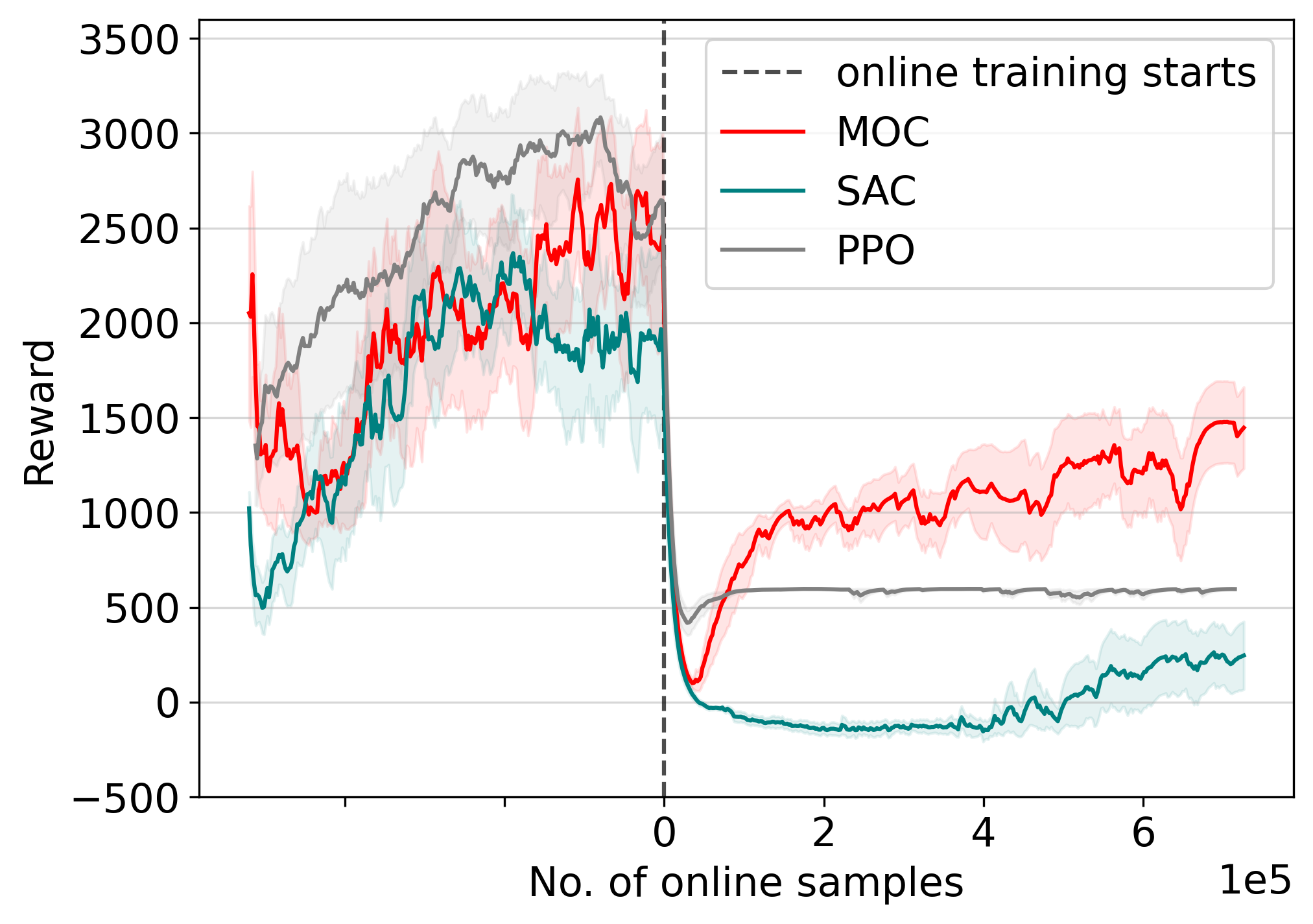}}
\subfigure[HalfCheetah Medium-Expert]{\includegraphics[width=0.23\textwidth]{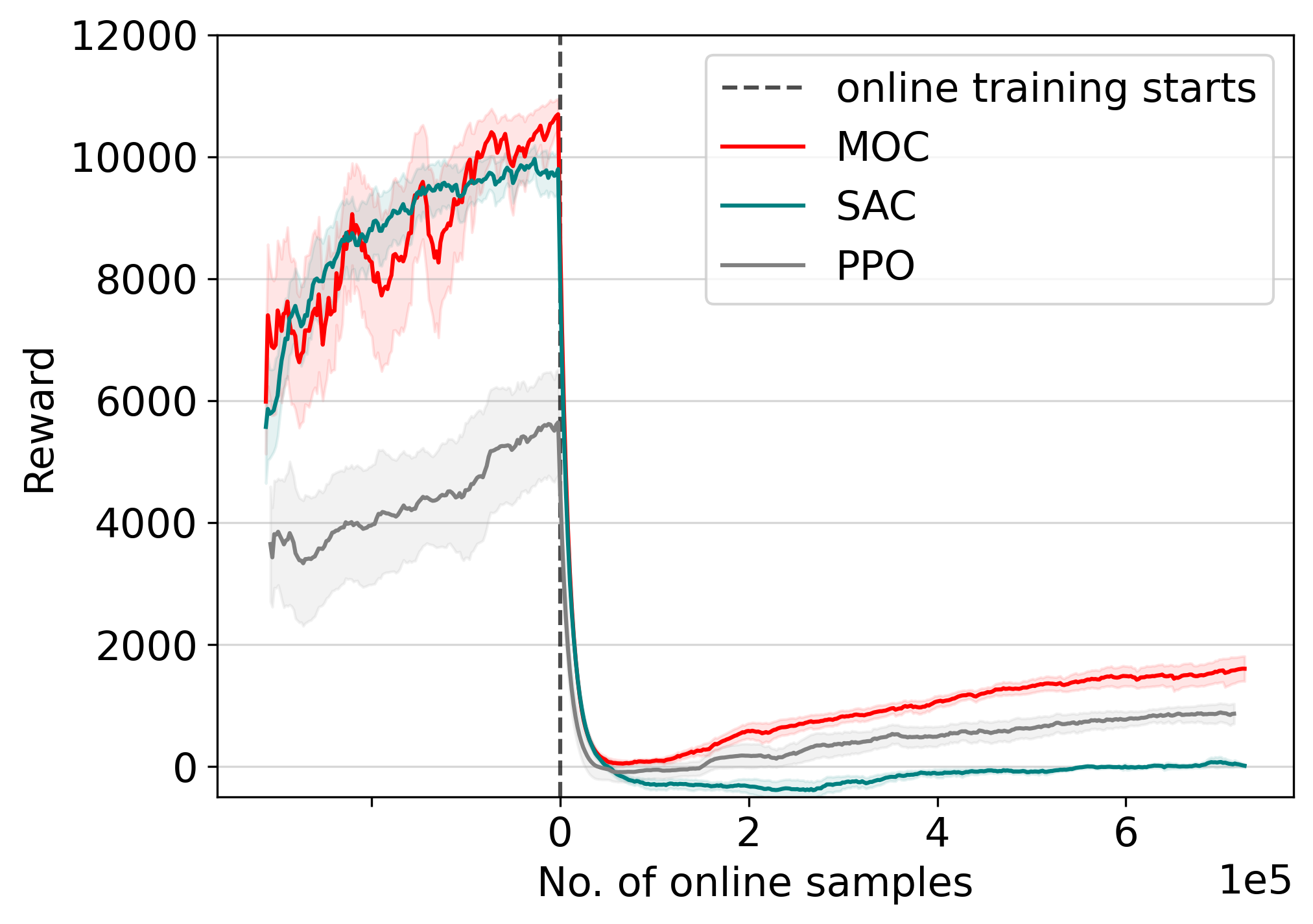}}
\caption{First row: results in Gym MuJoCo locomotion environments in the standard offline reinforcement learning setting. The captions specify the environment and the offline dataset on which the algorithms are learned. All the three algorithms are trained on the CVAE + P-MDP framework. Second row: results for the transfer task in Gym MuJoCo environments. The dip in the reward corresponds to the start of online training on the different tasks.}
\label{fig:standard}
\label{fig:transfer}
\end{figure*}

\begin{table*}[h]
    \centering
    \caption{Normalized scores on D4RL benchmark. Our algorithm scores are the running averages over five seeds. Results of MOReL, MOPO and CQL are reported from their respective papers.}
    \begin{tabular}{cccccc}
     \toprule
     Environment & Dataset & Offline MOC(Ours) & MOReL & MOPO & CQL \\
     \midrule
     Hopper & Medium & 76.8 & \textbf{95.4} & 28.0 & 86.6\\
     Hopper & Medium-Expert & 107.4 & 108.7 & 23.7 & \textbf{111} \\
     HalfCheetah & Medium & \textbf{45.04} & 42.1 & 42.3 & 44.4 \\
     HalfCheetah & Medium-Expert & \textbf{90.6} & 53.3 & 63.3 & 62.4 \\
    \bottomrule
    \end{tabular}
    \label{tab:normalized_score_locomotion}
\end{table*}

\subsection{Goal-conditioned setting}

In many problem settings, it is desirable to learn a general agent that can reach any goal given to it rather than retrain a different agent for each task that needs to be done. This has led to the study of algorithms that learn goal-conditioned policies that take the current state and intended goal as the input and output an action based on that.

In a hierarchical setting, the final intended goal is given to the high-level policy, which in turn creates a low-level sub-goal and passes it to the low-level policy to reach. The transitions in the offline dataset also contain the goal information.

To deal with such a scenario using our algorithm, we learn a goal-conditioned CVAE, in which both the encoder and decoder pass the goal along with the state as the conditioning input to learn the latent action space. The P-MDP model is left unchanged.

\section{EXPERIMENTS}
\label{sec:experiments}

\subsection{Standard setting}

For the standard setting, we consider the OpenAI Gym MuJoCo~\cite{Brockman2016OpenAIGym} continuous control environments, HalfCheetah-v2 and Hopper-v2. The agent controls a 2D figure, and the task is to make the figure move in the forward direction while remaining upright. We consider the "Medium" and "Medium-Expert" datasets in the D4RL benchmark~\cite{Fu2020D4RL} as the offline datasets on which our algorithm and baselines are run.

For the base online hierarchical algorithm, we use the Multi-updates Option Critic (MOC) algorithm~\cite{Klissarov2021FlexibleOptionLearning}. Here, a stochastic high-level policy chooses the option to execute, and the options are parameterized by neural networks with shared initial and hidden layers.

For the baselines, we consider Proximal Policy Optimization (PPO)~\cite{Schulman2017PPO} and Soft Actor-Critic (SAC)~\cite{Haarnoja2018SAC} algorithms run on top of our CVAE + P-MDP framework. This gives an insight into whether this framework preserves the advantages of using a hierarchical agent over a flat agent. Fig.~\ref{fig:standard} shows the results averaged over five runs, with the shaded region denoting the standard deviation.

Table~\ref{tab:normalized_score_locomotion} gives the results of our framework compared with other offline algorithms in the literature, which are Model-Based Offline Reinforcement Learning (MOReL)~\cite{Kidambi2020Morel}, Model-Based Offline Policy Optimization (MOPO)~\cite{Yu2020MOPO} and Conservative Q-Learning (CQL)~\cite{Kumar2020CQL}.

The performance of our algorithm is comparable to that of existing algorithms except on Hopper Medium. It outperforms the other methods in the HalfCheetah-v2 environment, and in the case of the HalfCheetah Medium-Expert dataset where MOReL attains a maximum unnormalized reward of around $8000$, it can be seen that the additional CVAE results in our method obtaining rewards more than $10,000$.

\subsection{Transfer setting}

For the task transfer setting, we consider the same MuJoCo locomotion environments as before. The offline dataset is the same as before, but during the online fine-tuning, the task is changed to a new one wherein the agent is rewarded for going in the backward direction, as opposed to the forward direction in the original task.

In the standard setting, we see that the hierarchical agent using MOC is performing as well as SAC and even PPO in some cases. However, when the task is switched, and the trained agents are updated to learn a new task online, we see in Fig.~\ref{fig:transfer} that MOC clearly outperforms the baselines. This shows that learning a hierarchy results in better sample complexity when the agent is asked to transfer its skills from one task to another. This also means that our framework preserves the advantages of a hierarchical agent over flat agents in the offline setting.

\begin{figure*}[t]
\centering
\vspace{3mm}
\subfigure[Goal (i)]{\includegraphics[width=0.30\textwidth]{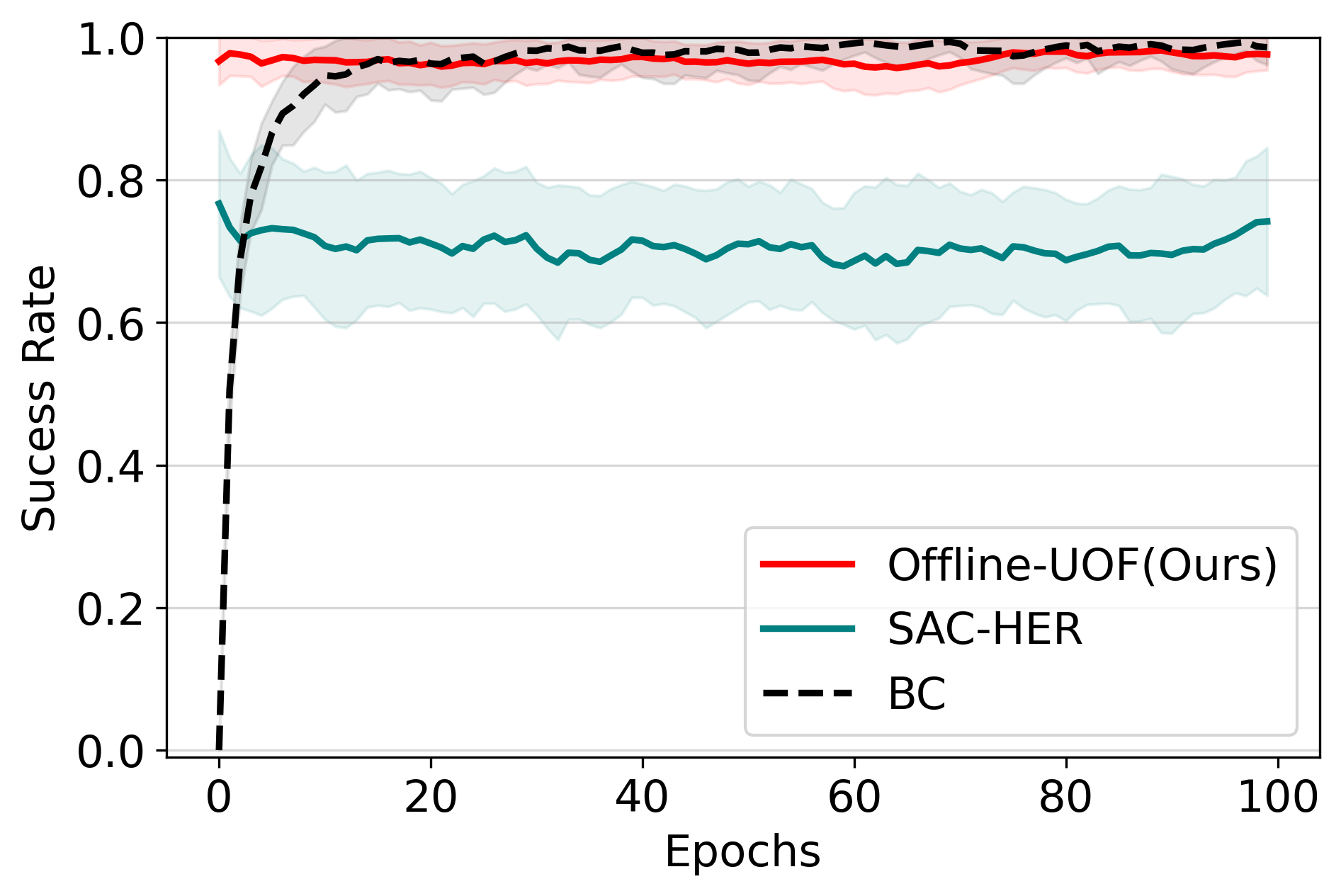}}
\subfigure[Goal (ii)]{\includegraphics[width=0.30\textwidth]{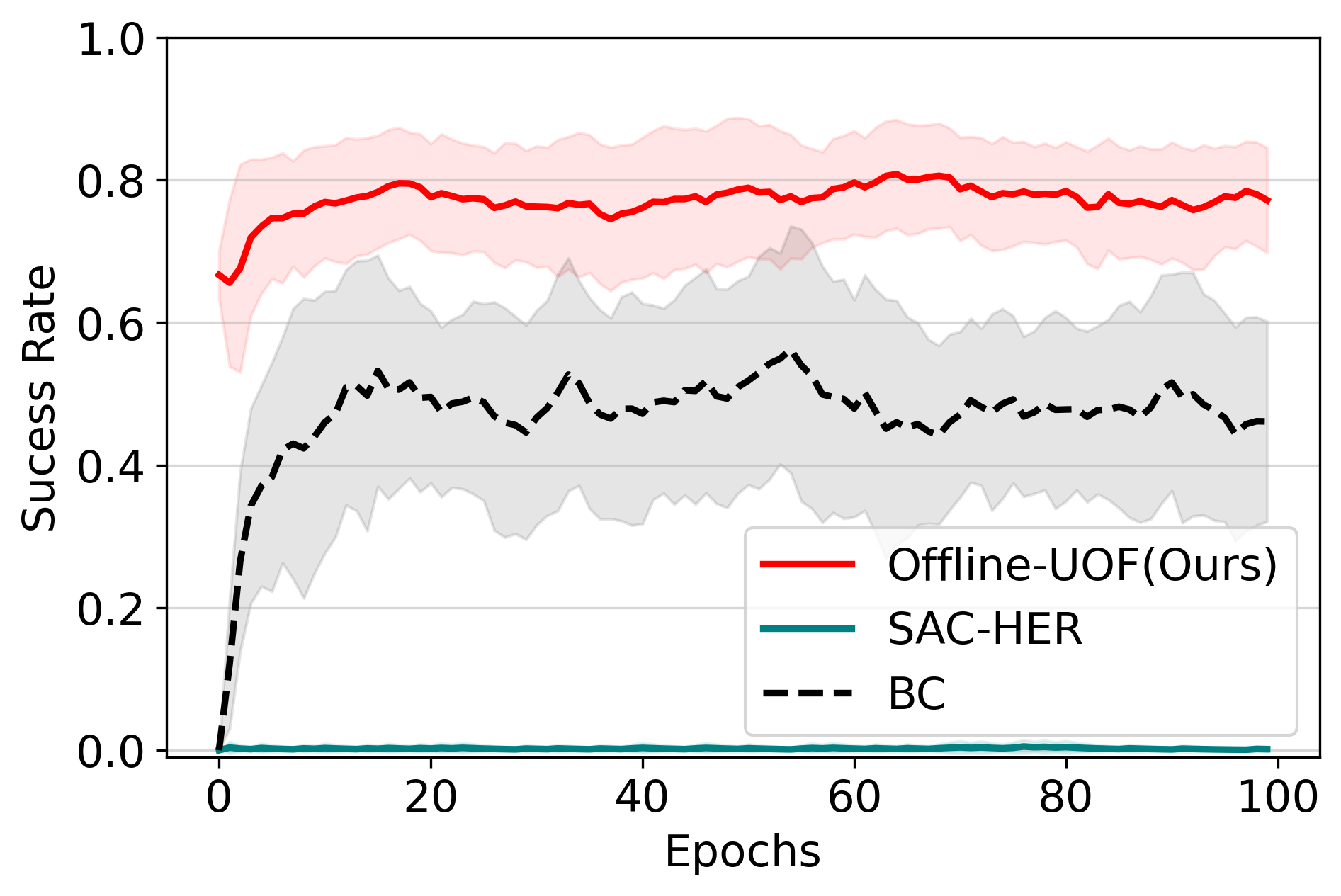}}
\subfigure[Goal (iii)]{\includegraphics[width=0.30\textwidth]{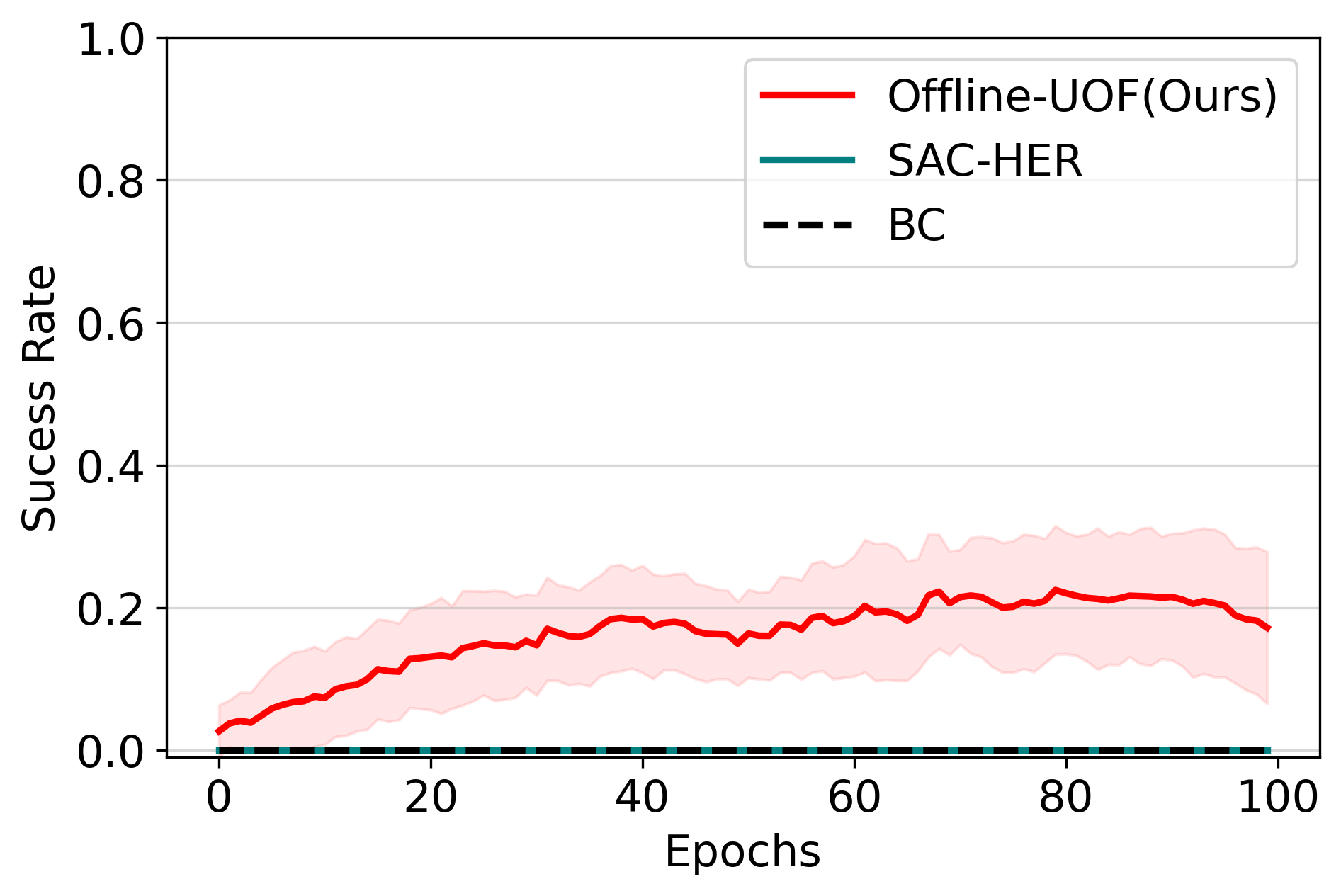}}
\subfigure[Goal (i)]{\includegraphics[width=0.30\textwidth]{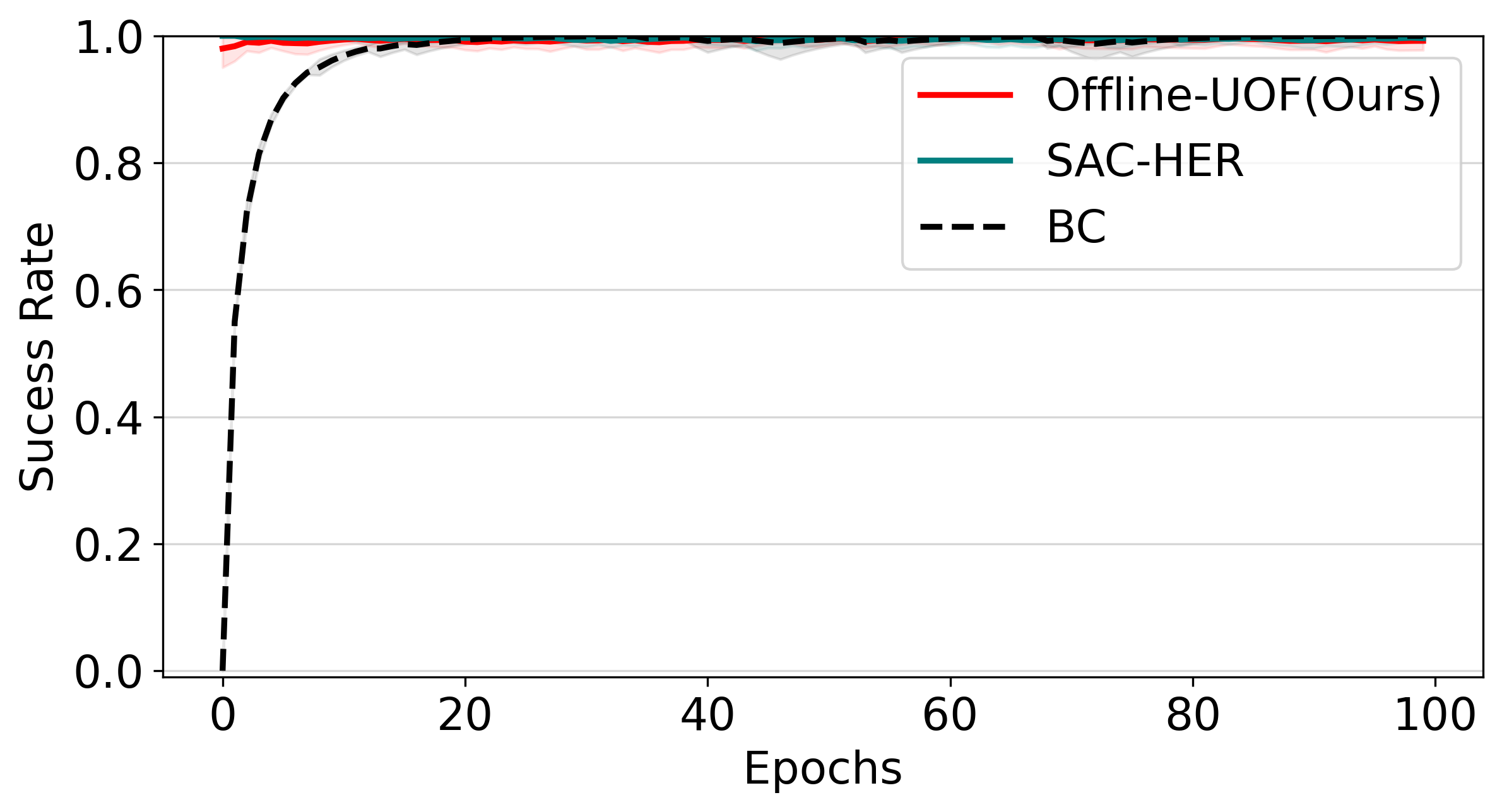}}
\subfigure[Goal (ii)]{\includegraphics[width=0.30\textwidth]{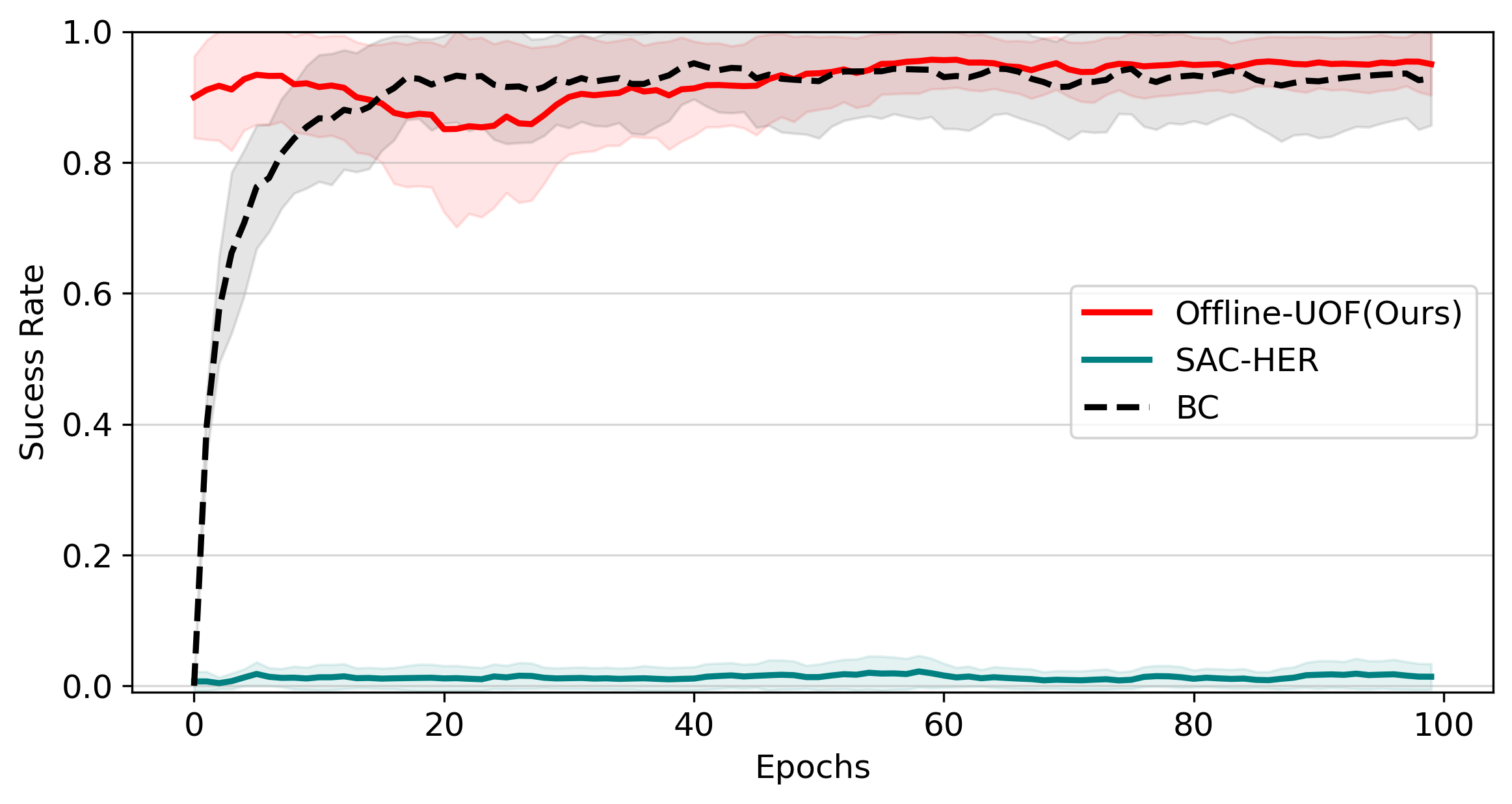}}
\subfigure[Goal (iii)]{\includegraphics[width=0.30\textwidth]{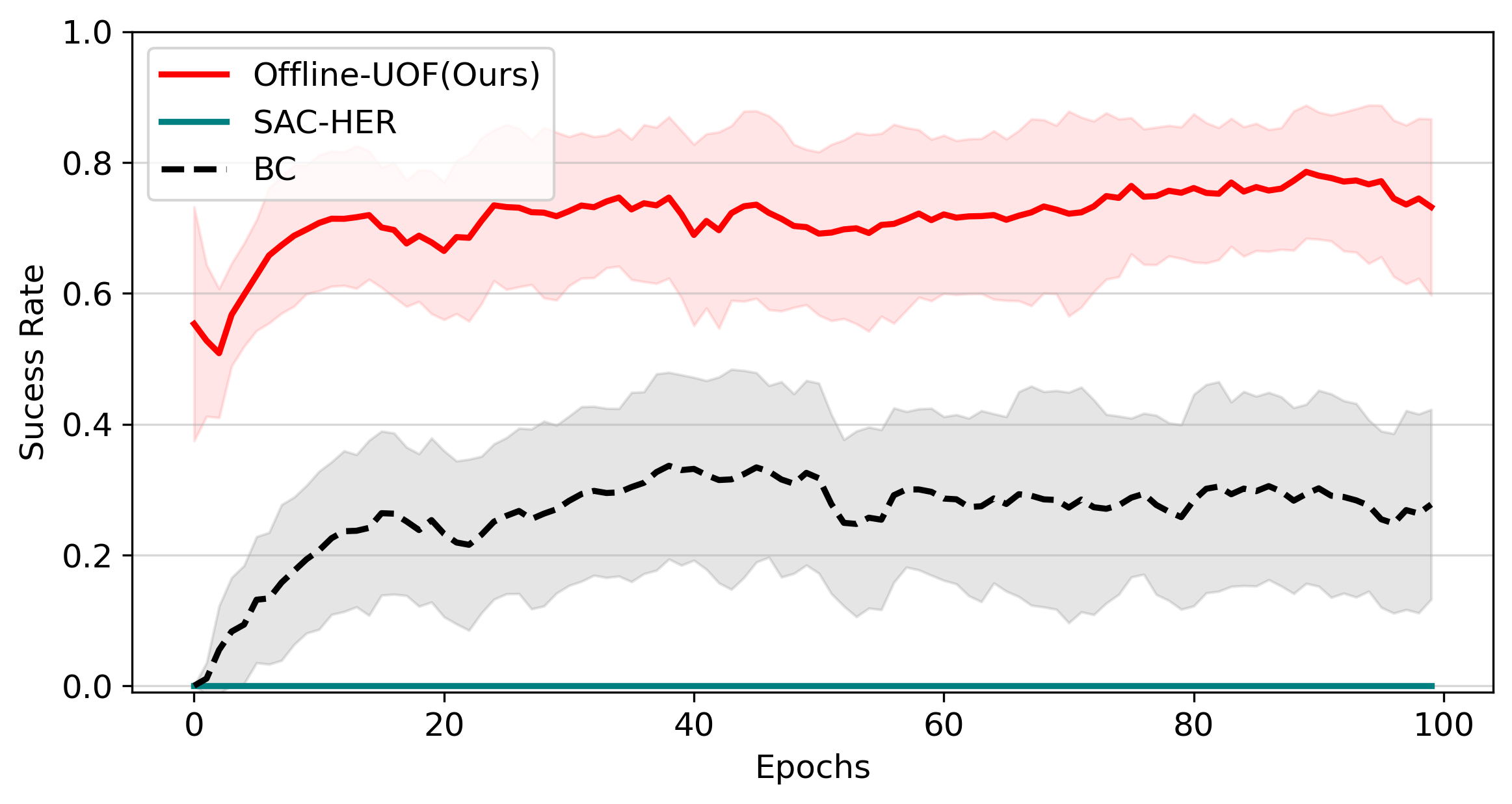}}
\caption{Results on the robotic-gripper block-stacking task for UOF along with SAC-HER and BC baselines, trained on the CVAE + P-MDP framework. The first and second rows depict the performance when trained on the Medium and Medium-Expert datasets respectively. Each plot shows the fraction of times the agents were able to reach the corresponding goal.}
\label{fig:goal_conditioned_setting}
\end{figure*}

\begin{table*}[h]
    \centering
    \caption{Performance on the second robotic block stacking task involving six goals.}
    \begin{tabular}{ccccccc}
     \toprule
     Algorithm & Goal (i) & (ii) & (iii) & (iv) & (v) & (vi) \\
     \midrule
     Our algorithm & 0.90 & \textbf{0.71} & \textbf{0.41} & 0.93 & \textbf{0.81} & \textbf{0.46}\\
     SAC+HER & 0.33 & 0.03 & 0 & 0.20 & 0.01 & 0\\
     BC & \textbf{0.99} & 0.50 & 0 & \textbf{0.99} & 0.20 & 0\\
    \bottomrule
    \end{tabular}
    \label{tab:robotics_task2}
\end{table*}

\subsection{Goal-conditioned setting}
\label{subsec:experiments:goal-conditioned}

For the goal-conditioned setting, we consider two block-stacking tasks wherein the agent controls a robotic gripper and has to handle and manipulate colored block-shaped objects based on the goal given, as in \cite{Yang2021HierarchicalRLWithUniversalPolicies}.

In the first task, the agent has to learn to achieve the three following goals: (i) grasping the blue block, (ii) placing it on the red block, and (iii) returning the gripper to the original position after successfully completing the previous tasks. It is given a $0$ reward when the goal is reached, $-1$ otherwise. The observations given to the agent consist of information about the positions and orientations of two blocks, and the goals are in the form of one-hot vectors. 

Similarly, the second task involves three blocks and six goals, with goals (i)-(iii) involving setting the blue block on the red block and goals (iv)-(vi) for setting the green block on the red block.

For these tasks, we consider the Universal Options Framework (UOF) algorithm that learns a goal-conditioned high-level policy and a goal-conditioned low-level policy. The output of the high-level policy is the low-level goal given to the low-level policy. The same sparse $\{ -1, 0 \}$ reward is given to the low-level policy as the low-level reward, depending on whether it reaches the low-level goal.

A fixed number of low-level goals are defined a priori as functions of the current state and correspond to the high-level goals. These are represented by the absolute Cartesian coordinates of the block and the gripper, along with the gripper width. The high-level policy chooses one of these as its high-level action and passes it to the low-level policy.

\begin{figure*}[t]
\centering
\vspace{3mm}
\subfigure[Goal (i)]{\includegraphics[width=0.30\textwidth]{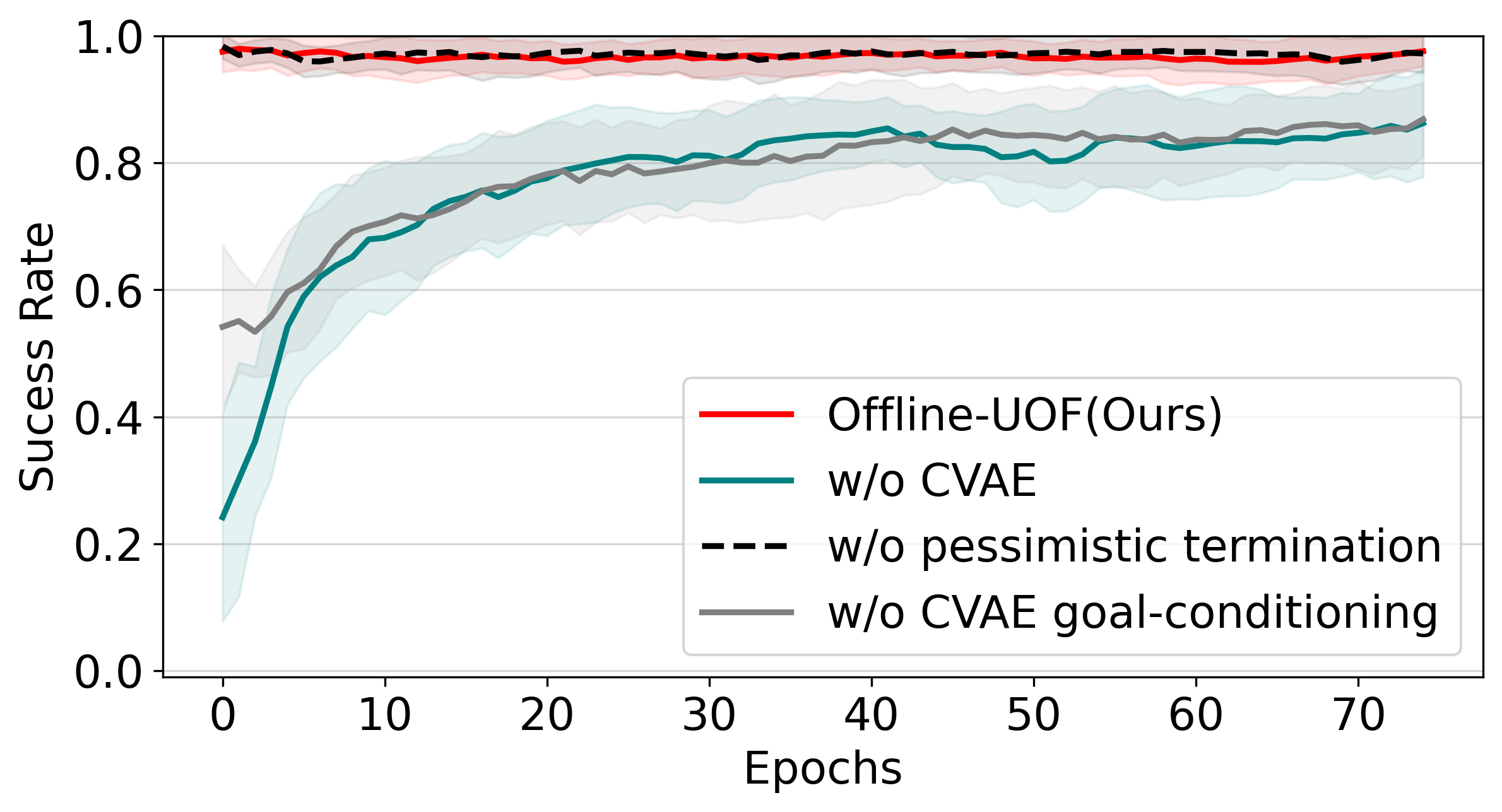} \label{fig:ablations:cvae}}
\subfigure[Goal (ii)]{\includegraphics[width=0.30\textwidth]{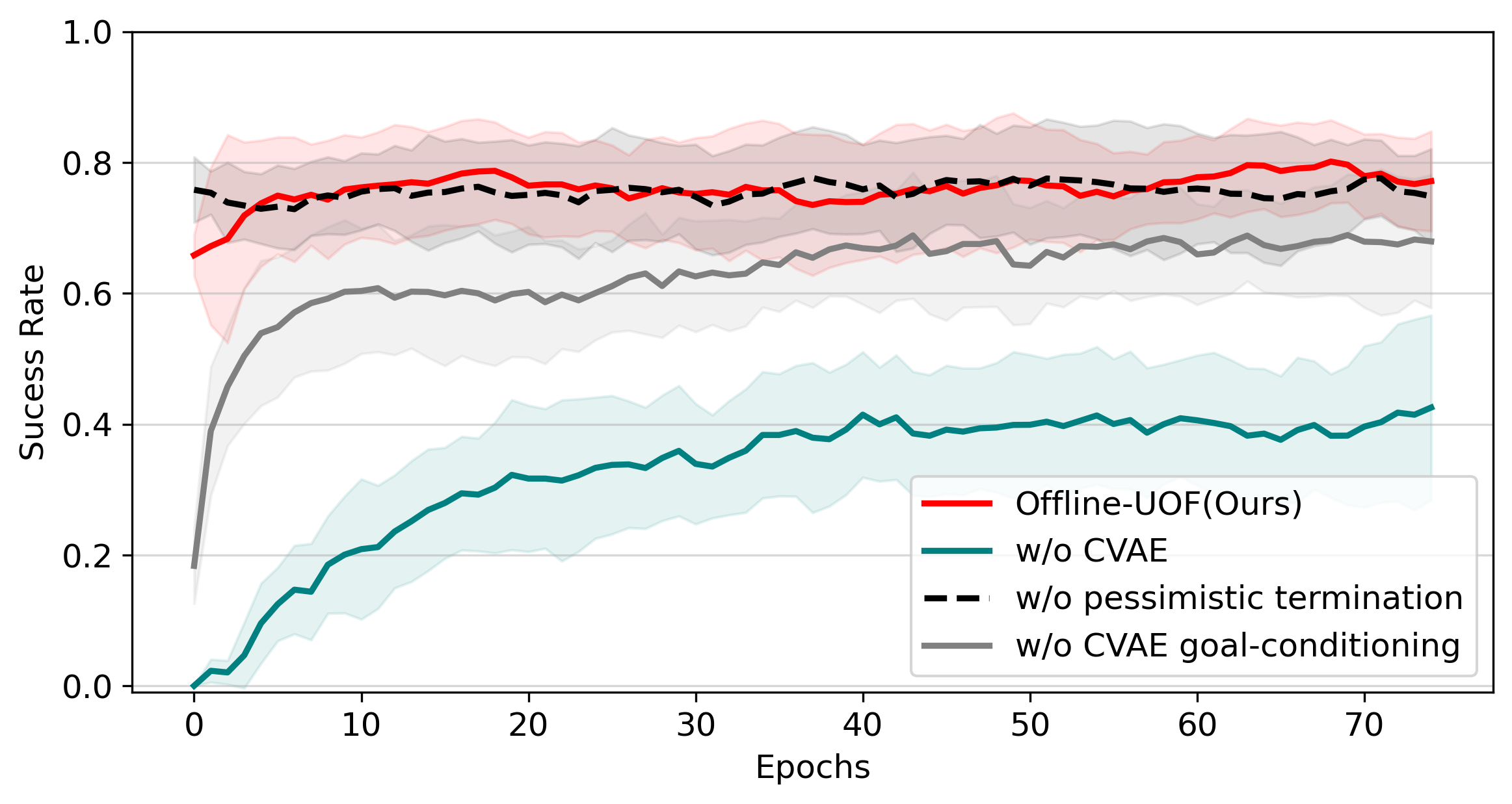} \label{fig:ablations:cvae_goal_conditioning}}
\subfigure[Goal (iii)]{\includegraphics[width=0.30\textwidth]{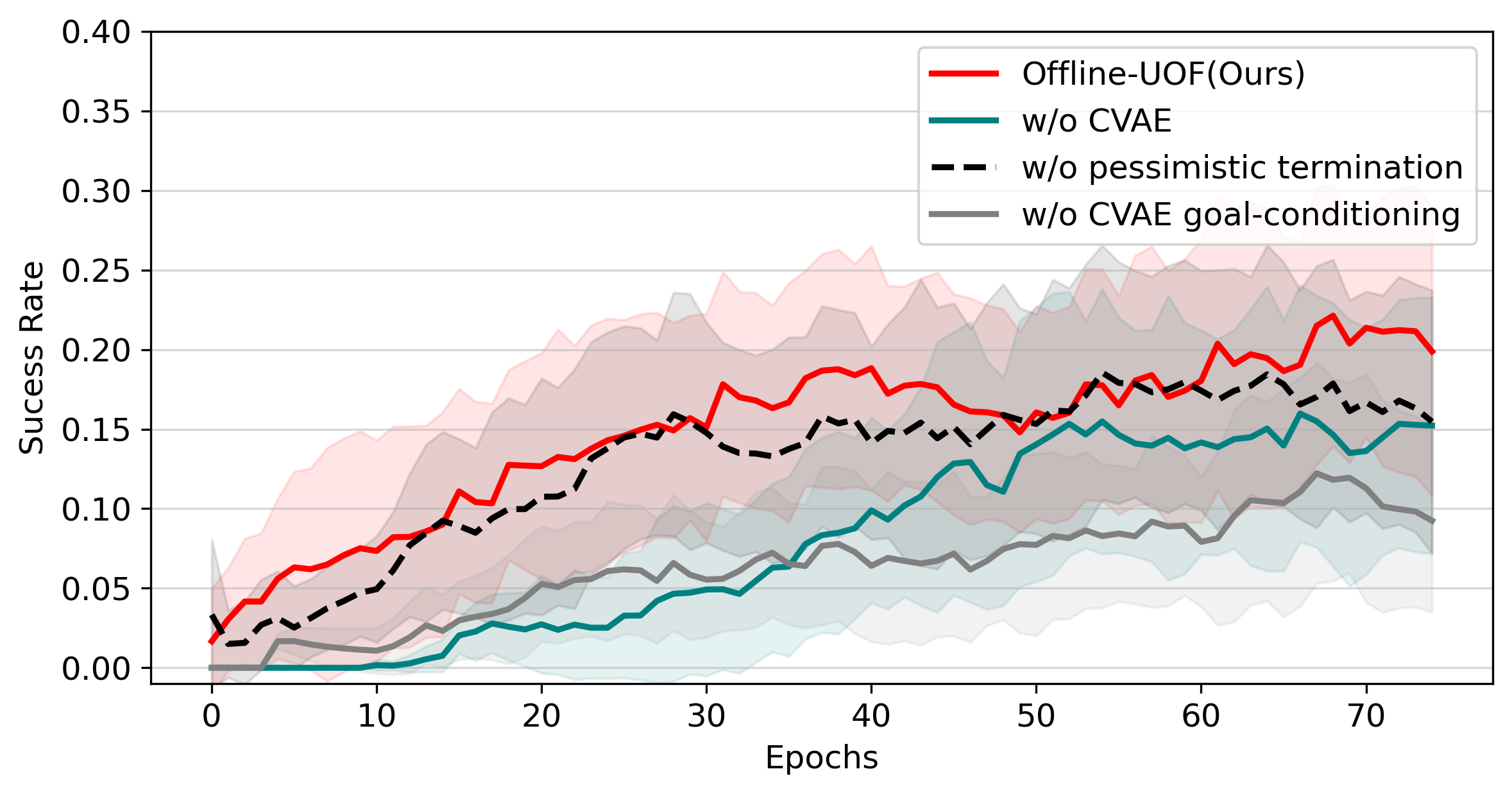}}
%\hspace{0.32\textwidth}
\subfigure[Ablation of pessimistic termination on Hopper Medium.]{\includegraphics[width=0.28\textwidth]{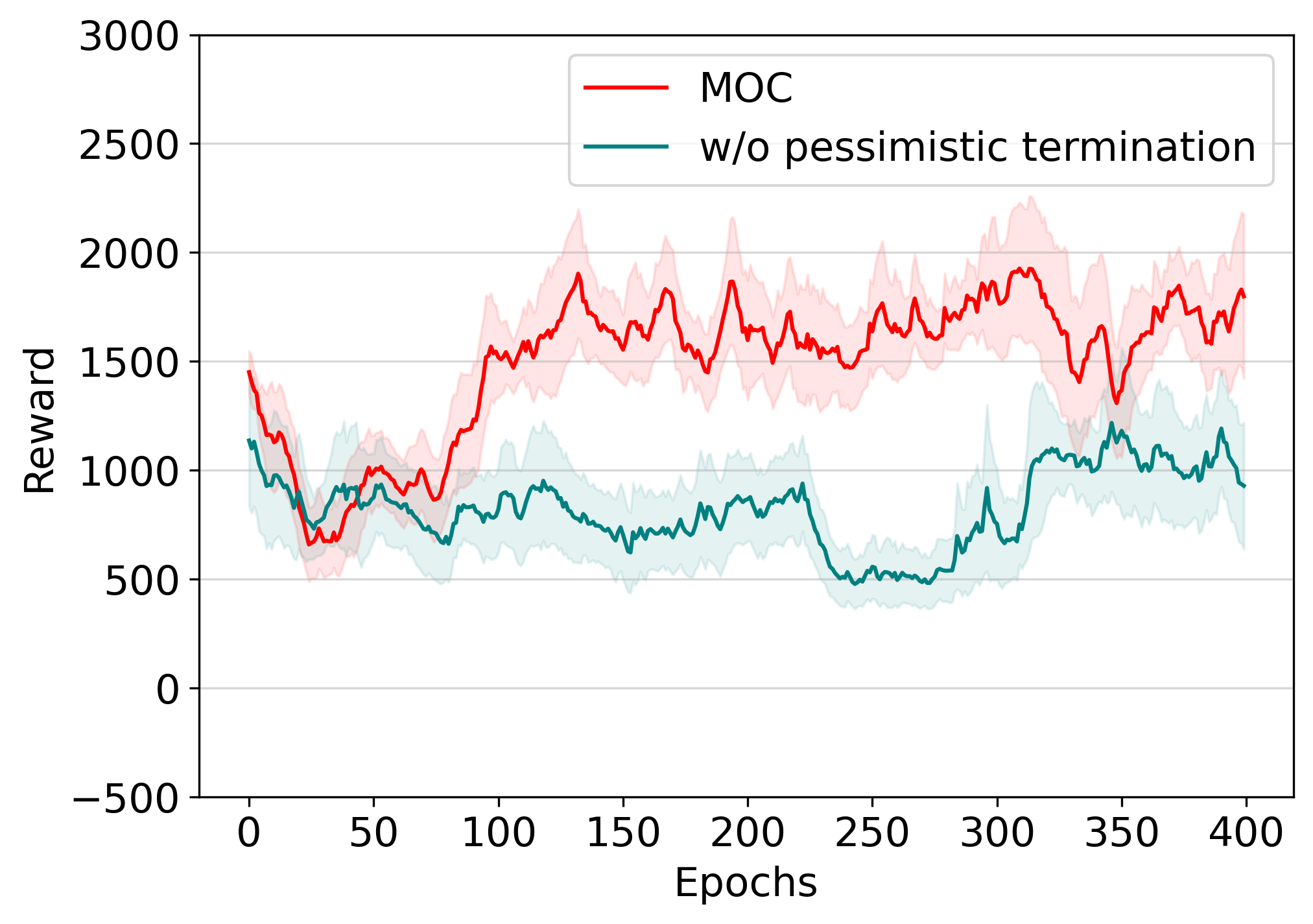}\label{fig:ablations:termination}}
\subfigure[Ablation of CVAE for Hopper task switch.]{\includegraphics[width=0.28\textwidth]{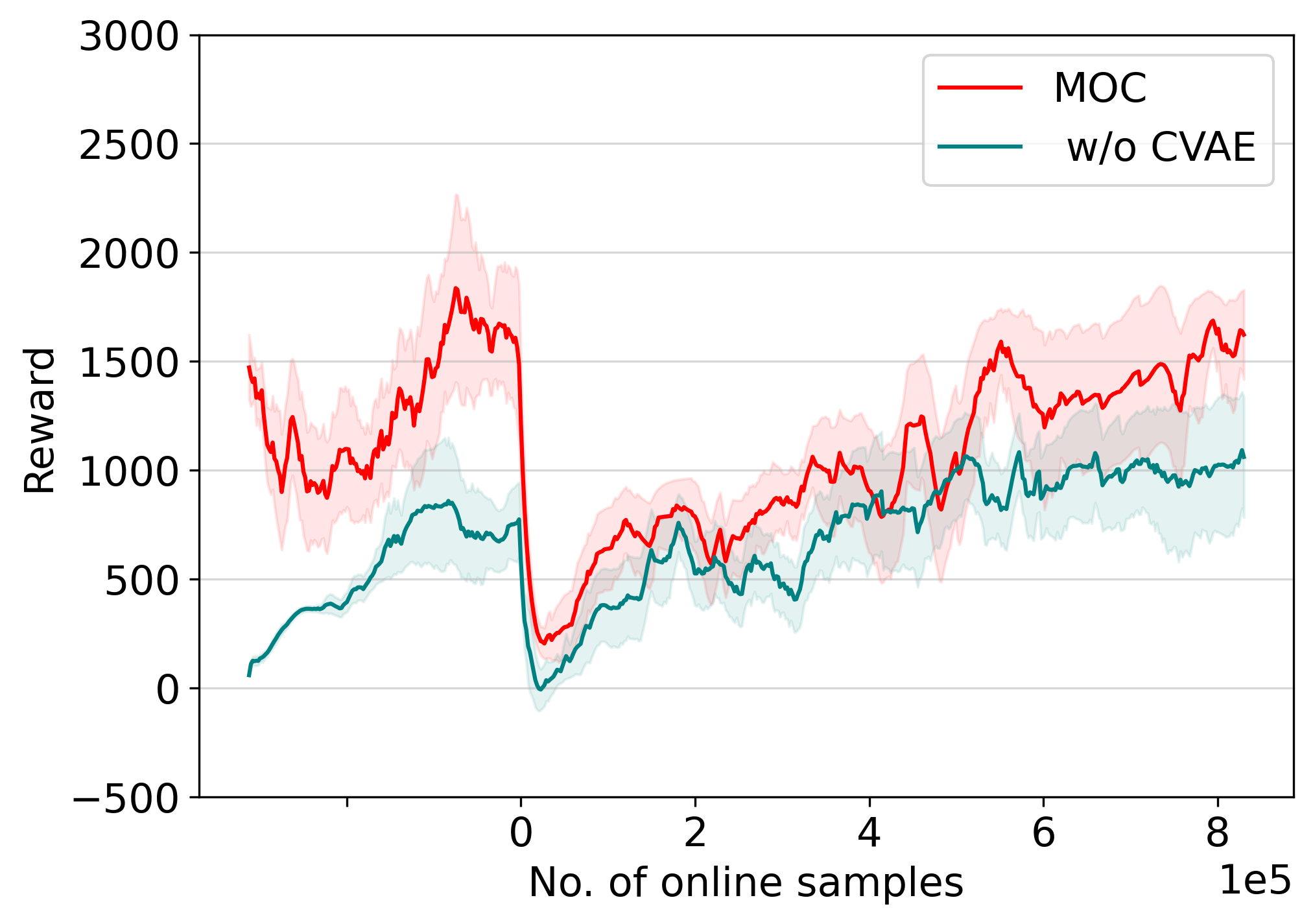}\label{fig:ablations:cvae_transfer}}
\caption{Results for the ablation experiments. The first row shows the ablation results of the UOF algorithm in the block-stacking task with and without a CVAE. The second row shows the results of the ablation of the pessimistic termination and the CVAE in the Hopper environment with the Medium offline dataset.}
\label{fig:ablations}
\end{figure*}

The high-level policy is trained using DIOL~\cite{Yang2021HierarchicalRLWithUniversalPolicies}, a deep-learning based goal-conditioned version of Intra-Option Learning~\cite{Sutton1998IOL}, and provided with abstract demonstrations, which are sequences of high-level actions to take to achieve the desired high-level goal. The low-level policy is learned using Deep Deterministic Policy Gradient (DDPG)~\cite{Lillicrap2016DDPG} and Hindsight Experience Replay (HER)~\cite{Andrychowicz2017HER} to deal with the sparse low-level rewards.

For this setting, we consider offline datasets consisting of goal-conditioned transitions of the form $(s_i, a_i, g_i, s_{i+1}, r_i, d_i)$, where $g_i$ is a low-level goal of the agent collecting the dataset. For the first task, we create two offline datasets, a ``Medium'' and a ``Medium-Expert'' dataset, of $1$M samples each for our experiments by running the UOF algorithm online for 20 and 50 epochs, respectively, and collecting samples by using these partially trained models to interact with the environment, and one dataset for the second task from a policy trained online for $200$ epochs.

We consider the SAC algorithm with HER as a baseline, which uses the low-level goals to learn goal-conditioned policies in the latent space of the goal-conditioned CVAE. We also consider a behavior cloning baseline that learns the behavior policy as a function of the current state and low-level goal.

For evaluation, each agent is given a goal to reach. For UOF, the goal is in the form of a one-hot vector. For the baselines, the goal is in the same format as the low-level goals of UOF. The results are shown in Fig.~\ref{fig:goal_conditioned_setting}. It is clear that while all agents can reach the initial goals, the baselines struggle to reach later goals due to a lack of planning, unlike the hierarchical algorithm.

Due to better performance of the behaviour policy on the Medium-Expert dataset, the behaviour cloning baseline gives almost perfect success rates for goals (i) and (ii), but fails to reliably reach goal (iii), whereas the UOF algorithm on top of CVAE + P-MDP gives around $0.8$ success rate for the final goal, surpassing both behaviour-cloning and SAC-HER.

\subsection{Options learned}

Due to the way the goals are defined in the block stacking task, reaching a goal requires reaching a sequence of previous goals in order. In the first task, the three possible high-level actions of the agent are defined corresponding to the three goals of the environment. So if the options are learned properly as intended, given a high-level goal, the agent should choose each of the previous high-level actions until the corresponding high-level goal is reached, and move on to the next high-level goal, and so on. For example, for reaching goal (ii), the agent should choose high-level action 0 to reach goal (i), then choose action 1 to reach goal (ii), and then choose action 2 to reach the intended goal.

\begin{figure*}[t]
\centering
\subfigure[Goal (i)]{\includegraphics[width=0.32\textwidth]{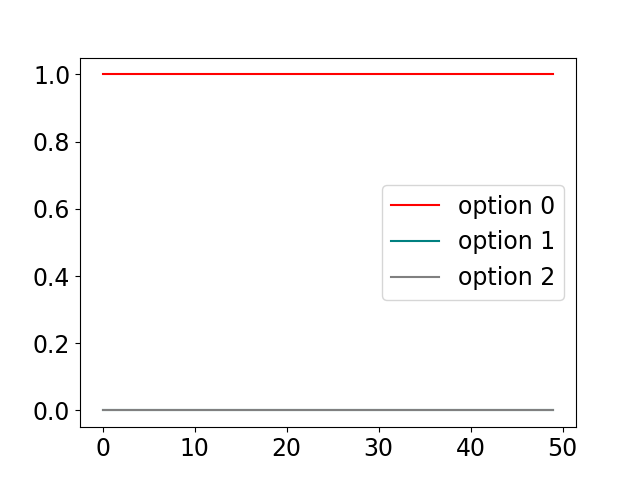}}
\subfigure[Goal (ii)]{\includegraphics[width=0.32\textwidth]{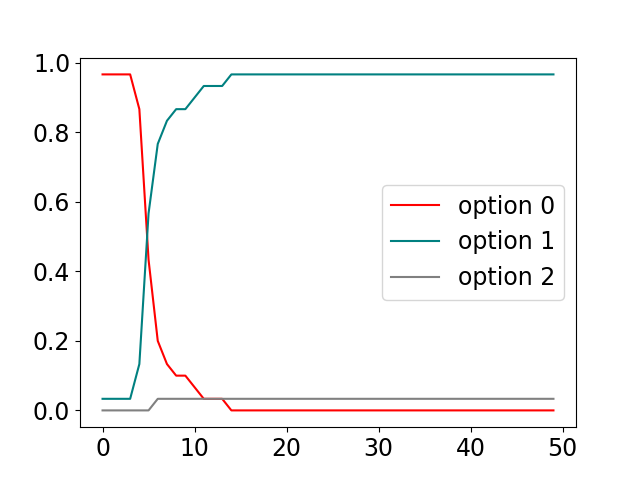}}
\subfigure[Goal (iii)]{\includegraphics[width=0.32\textwidth]{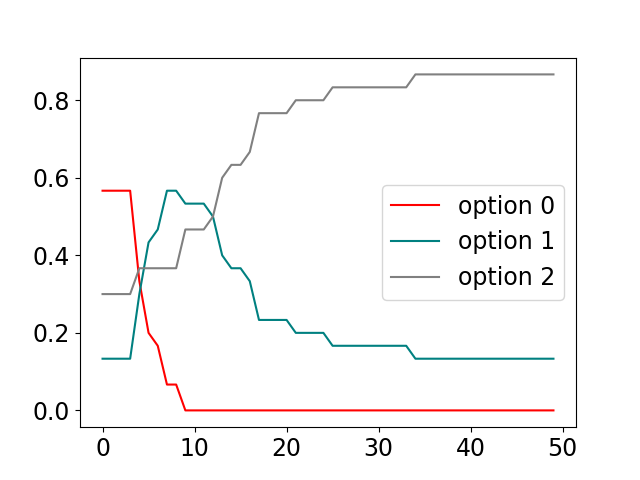}}
\caption{Fraction of times each high-level action is chosen by the trained UOF agent at each time step after being trained using Medium dataset.}
\label{fig:goal_conditioned_setting_options}
\end{figure*}

\begin{figure*}[t]
\centering
\subfigure[Goal (i)]{\includegraphics[width=0.32\textwidth]{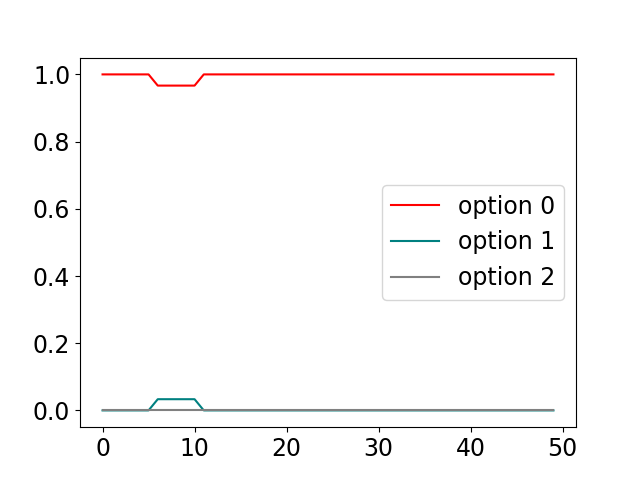}}
\subfigure[Goal (ii)]{\includegraphics[width=0.32\textwidth]{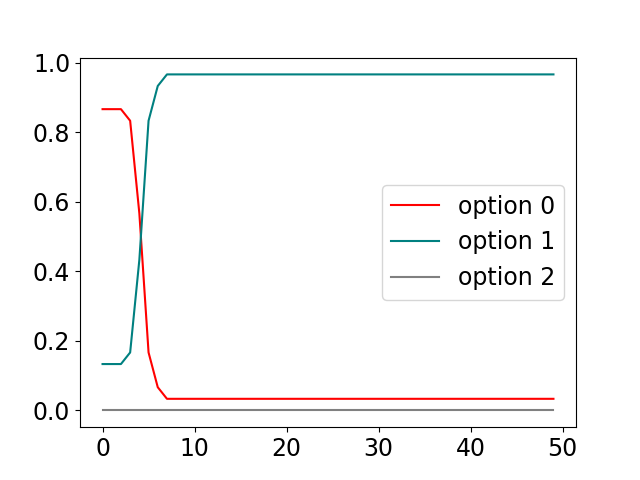}}
\subfigure[Goal (iii)]{\includegraphics[width=0.32\textwidth]{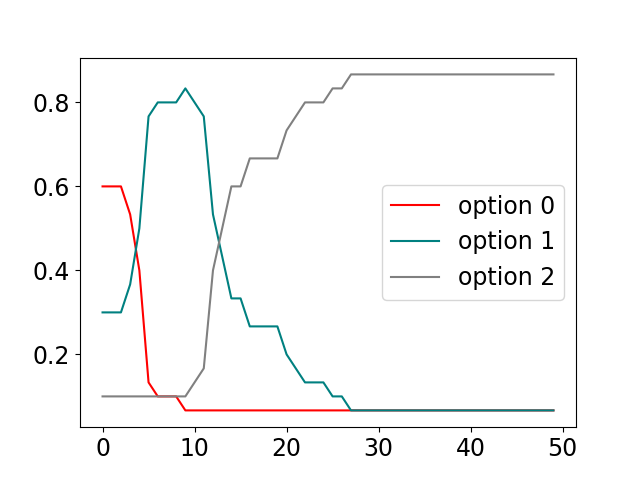}}
\caption{Fraction of times each high-level action is chosen by the trained UOF agent at each time step after being trained using Medium-Expert dataset.}
\label{fig:goal_conditioned_setting_options_medium_expert}
\end{figure*}

It turns out that this is actually the case for the agent learned using UOF on the offline dataset. This can be seen in Figures~\ref{fig:goal_conditioned_setting_options} and \ref{fig:goal_conditioned_setting_options_medium_expert}, which show the fraction of times a certain high-level action is chosen by the agent at a particular time step, when trained on the Medium and Medium-Expert datasets respectively till convergence and evaluated 30 times. The x-axis shows the time passed and the y-axis shows the fraction of times each option is active at that time step.

In Figure~\ref{fig:goal_conditioned_setting_options}, for goal $2$, it can be seen that option $0$ is chosen initially. Then the likelihood of choosing option $1$ peaks, followed by option $2$ to finally reach the goal. This order of choosing the options becomes even more evident and consistent in Figure~\ref{fig:goal_conditioned_setting_options_medium_expert} since the training has been done on a better dataset.

%The average success rate of goal2 being 0.833 is higher than that of in Medium dataset where the average success rate is around 0.2. From Figure~\ref{fig:goal_conditioned_setting_options_medium_expert}c), we can observe, there is a chronological order of choosing the options which is not evident in Figure~\ref{fig:goal_conditioned_setting_options} c).

\subsection{Ablations}

The necessity and effectiveness of the CVAE is a valid question since the MOReL algorithm does not need a CVAE to train an agent with a P-MDP. There might be an additional concern that limiting the actions that can be taken by the agent to the distribution learned by the CVAE might hinder generalization. However, we find that this is not the case, as seen in Fig.~\ref{fig:ablations}, from which it is clear that the CVAE not only enables learning a better policy but enables faster convergence as well, since the decoder acts as an initial pre-trained policy from which the agent can start and improve. The same can also be observed in the transfer setting, as seen in Fig.~\ref{fig:ablations:cvae_transfer} that shows the performance in the Hopper transfer task after training offline on the Medium dataset.

Another aspect of the algorithm in the goal-conditioned setting is the goal-conditioning of the CVAE itself. The same latent action taken by the low-level policy results in different actual actions executed in the environment based on the desired goal of the low-level policy and the current state. To study the utility of this approach, we conducted an ablation to compare the performance with that of an agent in which the CVAE is conditioned only on the state. It can be seen from Fig.~\ref{fig:ablations} that removing the goal-conditioning of the CVAE while retaining goal-conditioned high and low-level policies still results in a worse agent.

Additionally, we also conducted experiments without the pessimistic termination of the MDP when an uncertain state is encountered. We see that this component does not affect the performance significantly, possibly because all the episodes are restricted to only 25 time steps for this task and so early termination does not significantly affect the behavior of the learned policies. In environments with longer trajectories, pessimistic termination is an important component in the offline setting, as seen in Fig.~\ref{fig:ablations:termination}, which shows the results of this ablation in the Hopper-v2 environment trained on the Hopper Medium dataset.

\section{RELATED WORK}

The concept of options to incorporate temporal abstraction into reinforcement learning agents was introduced by \cite{Sutton1999BetweenMDPsAndSemiMDPs}. They studied this setting in the framework of Semi-Markov Decision Processes (SMDPs), in which each option is considered as an extended action that takes a variable amount of time steps to execute. The Option-Critic (OC) architecture~\cite{Bacon2017OptionCriticArchitecture} is a method to discover and learn options that are parameterized by deep neural networks and learned using policy gradient updates.

Over the years, many variants of the Option-Critic have been studied that aim to improve certain aspects of the algorithm. Interest Option Critic~\cite{Khetarpal2020OptionsOfInterest} generalizes initiation sets to learnable interest functions. Residual Soft Option Critic~\cite{Zhu2021EmpoweringDiversityAndIndividualityResidualSoftOptionCritic} extends OC by incorporating rewards based on the mutual information between different options and an entropy term. \cite{Smith2018InferenceBasedPolicyGradient} consider the options that gave rise to a trajectory as latent variables and update all the options to maximize the reward.

The use of a P-MDP for offline learning was introduced by \cite{Kidambi2020Morel}, who used a model-based Natural Policy Gradient (NPG) approach to train the agent in the P-MDP. Such ideas regarding pessimism or conservatism when dealing with out-of-distribution states or actions are incorporated by many offline algorithms in different ways. Model-based Offline Policy Optimization (MOPO)~\cite{Yu2020MOPO} also trains an arbitrary RL algorithm in an approximate MDP, penalizing the agent based on the aleatoric uncertainty of the current model at every time step. \cite{Lu2022RevistingDesignChoicesInOfflineModelBasedRL} empirically compares different choices of uncertainty penalization along with other design choices in model-based offline RL. Model-Based Offline Options (MO2)~\cite{Salter2022MO2} learns an option-transition model that predicts termination state distribution from a state-option-action tuple to optimize behavior cloning and predictability objectives.

The use of a CVAE trained on the offline dataset to implicitly constrain possible actions was introduced by \cite{Fujimoto2019BCQ} as a component of the Batch Constrained deep Q-learning (BCQ) algorithm. However, in BCQ, the actions output by the CVAE decoder are perturbed by a separate perturbation model learned using DDPG and the actions are selected in a model-free manner by sampling multiple actions and choosing the best based on the Q-values. The Policy in Latent Space (PLAS)~\cite{Zhou2021PLAS} algorithm is another model-free algorithm that trains an agent in the latent space of a CVAE and uses an optional perturbation layer after the decoder for out-of-distribution generalization. \cite{Nasiriany2019PlanningWithGoalConditionedPolicies} and \cite{Li2022HierarchicalPlanningThruGoalConditionedOfflineRL} use a VAE and CVAE, respectively to learn latent representations of the goal states in the dataset so as to sample meaningful goals in high-dimensional state spaces.

Learning hierarchical agents from offline data has been studied in the model-free setting. \cite{Ajay2021OPAL} learn a continuous space of low-level skills by treating a skill as a latent variable corresponding to a sub-trajectory in the offline dataset and the low-level policy as a function of the state and the skill.

\section{CONCLUSION}

In this work, we have proposed a framework to leverage online reinforcement learning algorithms for learning temporally abstracted policies from an offline dataset of transitions. We have validated our approach on a variety of continuous control tasks, such as MuJoCo locomotion tasks, as well as tasks involving planned manipulation of objects using a robotic gripper. We have shown that such an approach preserves the advantages of using a hierarchy of policies and performed ablation studies to demonstrate the necessity of every component of our algorithms.


%%%%%%%%%%%%%%%%%%%%%%%%%%%%%%%%%%%%%%%%%%%%%%%%%%%%%%%%%%%%%%%%%%%%%%%%%%%%%%%%

%%%%%%%%%%%%%%%%%%%%%%%%%%%%%%%%%%%%%%%%%%%%%%%%%%%%%%%%%%%%%%%%%%%%%%%%%%%%%%%%

\bibliographystyle{IEEEtran}
\bibliography{bibfile}

\begin{thebibliography}{10}
\providecommand{\url}[1]{#1}
\csname url@rmstyle\endcsname
\providecommand{\newblock}{\relax}
\providecommand{\bibinfo}[2]{#2}
\providecommand\BIBentrySTDinterwordspacing{\spaceskip=0pt\relax}
\providecommand\BIBentryALTinterwordstretchfactor{4}
\providecommand\BIBentryALTinterwordspacing{\spaceskip=\fontdimen2\font plus
\BIBentryALTinterwordstretchfactor\fontdimen3\font minus \fontdimen4\font\relax}
\providecommand\BIBforeignlanguage[2]{{%
\expandafter\ifx\csname l@#1\endcsname\relax
\typeout{** WARNING: IEEEtran.bst: No hyphenation pattern has been}%
\typeout{** loaded for the language `#1'. Using the pattern for}%
\typeout{** the default language instead.}%
\else
\language=\csname l@#1\endcsname
\fi
#2}}

\bibitem{Sutton1999BetweenMDPsAndSemiMDPs}
R.~S. Sutton, D.~Precup, and S.~Singh, ``Between {MDPs} and {semi-MDPs}: A framework for temporal abstraction in reinforcement learning,'' \emph{Artificial intelligence}, vol. 112, no. 1-2, pp. 181--211, 1999.

\bibitem{Klissarov2021FlexibleOptionLearning}
M.~Klissarov and D.~Precup, ``Flexible option learning,'' \emph{Advances in Neural Information Processing Systems}, vol.~34, pp. 4632--4646, 2021.

\bibitem{Bacon2017OptionCriticArchitecture}
P.-L. Bacon, J.~Harb, and D.~Precup, ``The option-critic architecture,'' in \emph{Proceedings of the AAAI conference on artificial intelligence}, vol.~31, no.~1, 2017.

\bibitem{Yang2021HierarchicalRLWithUniversalPolicies}
X.~Yang, Z.~Ji, J.~Wu, Y.-K. Lai, C.~Wei, G.~Liu, and R.~Setchi, ``Hierarchical reinforcement learning with universal policies for multistep robotic manipulation,'' \emph{IEEE Transactions on Neural Networks and Learning Systems}, vol.~33, no.~9, pp. 4727--4741, 2021.

\bibitem{Kidambi2020Morel}
R.~Kidambi, A.~Rajeswaran, P.~Netrapalli, and T.~Joachims, ``{MOReL:} model-based offline reinforcement learning,'' \emph{Advances in neural information processing systems}, vol.~33, pp. 21\,810--21\,823, 2020.

\bibitem{Brockman2016OpenAIGym}
G.~Brockman, V.~Cheung, L.~Pettersson, J.~Schneider, J.~Schulman, J.~Tang, and W.~Zaremba, ``Openai gym,'' 2016.

\bibitem{Fu2020D4RL}
J.~Fu, A.~Kumar, O.~Nachum, G.~Tucker, and S.~Levine, ``{D4RL:} datasets for deep data-driven reinforcement learning,'' \emph{arXiv preprint arXiv:2004.07219}, 2020.

\bibitem{Schulman2017PPO}
J.~Schulman, F.~Wolski, P.~Dhariwal, A.~Radford, and O.~Klimov, ``Proximal policy optimization algorithms,'' \emph{arXiv preprint arXiv:1707.06347}, 2017.

\bibitem{Haarnoja2018SAC}
T.~Haarnoja, A.~Zhou, P.~Abbeel, and S.~Levine, ``Soft actor-critic: Off-policy maximum entropy deep reinforcement learning with a stochastic actor,'' in \emph{International conference on machine learning}.\hskip 1em plus 0.5em minus 0.4em\relax PMLR, 2018, pp. 1861--1870.

\bibitem{Yu2020MOPO}
T.~Yu, G.~Thomas, L.~Yu, S.~Ermon, J.~Y. Zou, S.~Levine, C.~Finn, and T.~Ma, ``{MOPO:} model-based offline policy optimization,'' \emph{Advances in Neural Information Processing Systems}, vol.~33, pp. 14\,129--14\,142, 2020.

\bibitem{Kumar2020CQL}
A.~Kumar, A.~Zhou, G.~Tucker, and S.~Levine, ``Conservative q-learning for offline reinforcement learning,'' \emph{Advances in Neural Information Processing Systems}, vol.~33, pp. 1179--1191, 2020.

\bibitem{Sutton1998IOL}
R.~S. Sutton, D.~Precup, and S.~P. Singh, ``Intra-option learning about temporally abstract actions,'' in \emph{Proceedings of the Fifteenth International Conference on Machine Learning}, ser. ICML '98.\hskip 1em plus 0.5em minus 0.4em\relax San Francisco, CA, USA: Morgan Kaufmann Publishers Inc., 1998, p. 556–564.

\bibitem{Lillicrap2016DDPG}
\BIBentryALTinterwordspacing
T.~P. Lillicrap, J.~J. Hunt, A.~Pritzel, N.~Heess, T.~Erez, Y.~Tassa, D.~Silver, and D.~Wierstra, ``Continuous control with deep reinforcement learning,'' in \emph{4th International Conference on Learning Representations, {ICLR} 2016, San Juan, Puerto Rico, May 2-4, 2016, Conference Track Proceedings}, Y.~Bengio and Y.~LeCun, Eds., 2016. [Online]. Available: \url{http://arxiv.org/abs/1509.02971}
\BIBentrySTDinterwordspacing

\bibitem{Andrychowicz2017HER}
M.~Andrychowicz, F.~Wolski, A.~Ray, J.~Schneider, R.~Fong, P.~Welinder, B.~McGrew, J.~Tobin, O.~Pieter~Abbeel, and W.~Zaremba, ``Hindsight experience replay,'' \emph{Advances in neural information processing systems}, vol.~30, 2017.

\bibitem{Khetarpal2020OptionsOfInterest}
K.~Khetarpal, M.~Klissarov, M.~Chevalier-Boisvert, P.-L. Bacon, and D.~Precup, ``Options of interest: Temporal abstraction with interest functions,'' in \emph{Proceedings of the AAAI Conference on Artificial Intelligence}, vol.~34, no.~04, 2020, pp. 4444--4451.

\bibitem{Zhu2021EmpoweringDiversityAndIndividualityResidualSoftOptionCritic}
A.~Zhu, F.~Chen, H.~Xu, D.~Ouyang, and J.~Shao, ``Empowering the diversity and individuality of option: Residual soft option critic framework,'' \emph{IEEE Transactions on Neural Networks and Learning Systems}, 2021.

\bibitem{Smith2018InferenceBasedPolicyGradient}
M.~Smith, H.~Hoof, and J.~Pineau, ``An inference-based policy gradient method for learning options,'' in \emph{International Conference on Machine Learning}.\hskip 1em plus 0.5em minus 0.4em\relax PMLR, 2018, pp. 4703--4712.

\bibitem{Lu2022RevistingDesignChoicesInOfflineModelBasedRL}
\BIBentryALTinterwordspacing
C.~Lu, P.~J. Ball, J.~Parker{-}Holder, M.~A. Osborne, and S.~J. Roberts, ``Revisiting design choices in offline model based reinforcement learning,'' in \emph{The Tenth International Conference on Learning Representations, {ICLR} 2022, Virtual Event, April 25-29, 2022}.\hskip 1em plus 0.5em minus 0.4em\relax OpenReview.net, 2022. [Online]. Available: \url{https://openreview.net/forum?id=zz9hXVhf40}
\BIBentrySTDinterwordspacing

\bibitem{Salter2022MO2}
S.~Salter, M.~Wulfmeier, D.~Tirumala, N.~Heess, M.~Riedmiller, R.~Hadsell, and D.~Rao, ``{Mo2:} model-based offline options,'' in \emph{Conference on Lifelong Learning Agents}.\hskip 1em plus 0.5em minus 0.4em\relax PMLR, 2022, pp. 902--919.

\bibitem{Fujimoto2019BCQ}
S.~Fujimoto, D.~Meger, and D.~Precup, ``Off-policy deep reinforcement learning without exploration,'' in \emph{International conference on machine learning}.\hskip 1em plus 0.5em minus 0.4em\relax PMLR, 2019, pp. 2052--2062.

\bibitem{Zhou2021PLAS}
W.~Zhou, S.~Bajracharya, and D.~Held, ``Plas: Latent action space for offline reinforcement learning,'' in \emph{Conference on Robot Learning}.\hskip 1em plus 0.5em minus 0.4em\relax PMLR, 2021, pp. 1719--1735.

\bibitem{Nasiriany2019PlanningWithGoalConditionedPolicies}
S.~Nasiriany, V.~Pong, S.~Lin, and S.~Levine, ``Planning with goal-conditioned policies,'' \emph{Advances in Neural Information Processing Systems}, vol.~32, 2019.

\bibitem{Li2022HierarchicalPlanningThruGoalConditionedOfflineRL}
J.~Li, C.~Tang, M.~Tomizuka, and W.~Zhan, ``Hierarchical planning through goal-conditioned offline reinforcement learning,'' \emph{IEEE Robotics and Automation Letters}, vol.~7, no.~4, pp. 10\,216--10\,223, 2022.

\bibitem{Ajay2021OPAL}
\BIBentryALTinterwordspacing
A.~Ajay, A.~Kumar, P.~Agrawal, S.~Levine, and O.~Nachum, ``{OPAL:} offline primitive discovery for accelerating offline reinforcement learning,'' in \emph{9th International Conference on Learning Representations, {ICLR} 2021, Virtual Event, Austria, May 3-7, 2021}.\hskip 1em plus 0.5em minus 0.4em\relax OpenReview.net, 2021. [Online]. Available: \url{https://openreview.net/forum?id=V69LGwJ0lIN}
\BIBentrySTDinterwordspacing

\bibitem{Liaw2018Tune}
R.~Liaw, E.~Liang, R.~Nishihara, P.~Moritz, J.~E. Gonzalez, and I.~Stoica, ``Tune: A research platform for distributed model selection and training,'' \emph{arXiv preprint arXiv:1807.05118}, 2018.

\bibitem{stable-baselines}
A.~Hill, A.~Raffin, M.~Ernestus, A.~Gleave, A.~Kanervisto, R.~Traore, P.~Dhariwal, C.~Hesse, O.~Klimov, A.~Nichol, M.~Plappert, A.~Radford, J.~Schulman, S.~Sidor, and Y.~Wu, ``Stable baselines,'' \url{https://github.com/hill-a/stable-baselines}, 2018.

\end{thebibliography}

\appendix[Robotics environment]

We consider the first two block-stacking tasks described in \cite{Yang2021HierarchicalRLWithUniversalPolicies}, wherein the agent has to manipulate given colored blocks. The first task consists of three goals to (i) catch the blue block, (ii) set it on the red block, and (iii) return to the initial position. Similarly, the second task consists of six goals corresponding to placing a blue block or a green block on the red block.

\subsection{Dynamics Model}
The dynamics model for the robotics environment consists of just a transition model. The reward function is not learned but defined to be $0$ when the goal is reached and $-1$ otherwise. The transition model comprises a Multi-Layer Perceptron (MLP) with two hidden layers of $512$ ReLU activated nodes each. We use an ensemble of $5$ models to compute the discrepancy in the next state prediction. The states are normalized and the models predict the residual for next state. Since normalized states are used, the dynamics model is learned using the Adam optimizer to minimize the $l_1$ loss.

\subsection{CVAE architecture}

Our CVAE consists of encoder and decoder networks each of which comprises 2 hidden layers with $750$ ReLU activated nodes. The low-level goal is passed along with the state as input to the CVAE components. The latent dimension is same as that of the action dimension. Similar to the dynamics model, we use Adam optimizer and $l_1$ Loss. Once the CVAE is learned, the encoder is discarded and just the decoder part of the CVAE is used to convert the latent actions of the agent to the real actions to be taken in the P-MDP or the environment during training and evaluation of the RL agent respectively.

\subsection{Hyper-parameters}

For training the dynamics models and the CVAE, we use a train-validation split of $85:15$. The learning rate and the number of hidden neurons in the MLP are decided by a grid-search technique on the basis of the respective validation losses. The search is done using Ray Tune~\cite{Liaw2018Tune} to randomly select and evaluate 10 out of all possible combinations. The ranges on which the hyper-parameters are searched are given in Table~\ref{tab:hyperparams:robotics_dynamics_cvae_search}.

\begin{table}[h]
    \centering
    \caption{Grid search range for hyper-parameters of the transition models and CVAE networks.}
    \begin{tabular}{cc}
        \toprule
        Hyper-parameter & Search Range  \\
        \midrule
        Learning rate & $\{ 10^{-3}, 10^{-4}, 10^{-5}$, \\
        ~ & $2 \times 10^{-4}, 5 \times 10^{-4} \}$ \\
        Transition model hidden layer size & $\{ 128, 256, 512, 1024 \}$ \\
        CVAE hidden layer size &  $ \{ 256,512,720,1024 \} $ \\
        \bottomrule
    \end{tabular}\label{tab:hyperparams:robotics_dynamics_cvae_search}
\end{table}

\begin{table}[h]
    \centering
    \caption{Hyper-parameters for dynamics model and CVAE in the robotics environment.}
    \begin{tabular}{cc}
    \toprule
        Hyper-parameter & Value \\
    \midrule
        Dynamics Model & MLP$(512,512)$ \\
        Activation & ReLU \\
        Batch size & $256$ \\
        Learning rate & $10^{-4}$ \\
        Discrepancy quantile & $99$ \\
        Negative reward & $50$ \\
        No. of models in ensemble & $5$ \\
        ~ &  ~ \\
        CVAE & MLP$(720,720)$ \\
        Activation & ReLU \\
        Batch size & $128$ \\
        Learning rate & $10^{-4}$ \\
    \bottomrule
    \end{tabular}
    \label{tab:hyperparams:robotics}
\end{table}

At any state, when a specific action is taken, a discrepancy is calculated as the variance of the next state prediction by the transition functions of model ensemble. As described previously, when the discrepancy at a particular state-action pair exceeds a certain threshold, the episode is terminated with a high negative reward.

The threshold can be defined in one of two ways. Once the ensemble is trained and fixed, the discrepancy value is calculated for each state-action pairs in the offline dataset. Using these values, the threshold can be calculated as either a quantile of these values, or a fraction of the maximum discrepancy value on the dataset.

Which of the above two methods to use, and the corresponding quantile or fraction value, are both hyper-parameters, as well as the reward penalty given when the discrepancy exceeds this threshold and the episode is terminated.

The discrepancy threshold is set by evaluating the UOF agent in the original environment using the CVAE. For this environment, we set the threshold as a quantile of the discrepancy values induced by the ensemble on the offline dataset. The final hyper-parameters are given in Table~\ref{tab:hyperparams:robotics}. The hyper-parameters for SAC-HER are the default hyper-parameters in Stable Baselines 3.

The SAC-HER algorithm is taken from Stable-Baselines3 \cite{stable-baselines} and the UOF code has been adopted from its official implementation~\cite{Yang2021HierarchicalRLWithUniversalPolicies}.

\appendix[Locomotion environments]

For the locomotion environments, we consider two MuJoCo continuous control tasks, Hopper-v2 and HalfCheetah-v2, in OpenAI Gym~\cite{Brockman2016OpenAIGym}. The agent controls a 2D figure and the task is to make the figure move in the forward direction while remaining upright. We consider the ``Medium'' and ``Medium-Expert'' datasets in the D4RL benchmark~\cite{Fu2020D4RL} as the offline datasets on which our algorithm and baselines are run. The Medium dataset comprises 1M transitions and the Medium-Expert dataset comprises 2M. After transfer, the Gym environment is modified such that the environment gives reward when the agent moves in the backward direction.

\subsection{Dynamics Model}
Our dynamics model consists of a transition model as well as a reward model, each of which comprises an MLP with two hidden layers of 512 ReLU activated nodes. We use an ensemble of 5 models to compute the discrepancy in the next state prediction. The dynamics model and the reward model are both learned using the Adam optimizer and the $l_1$ Loss.

\begin{table}[h]
    \centering
    \caption{Grid search range for hyper-parameters of the transition models and CVAE networks.}
    \begin{tabular}{cc}
        \toprule
        Hyper-parameter & Search Range  \\
        \midrule
        Learning rate & $\{ 10^{-3}, 10^{-4}, 10^{-5}$, \\
        ~ & $2 \times 10^{-4}, 5 \times 10^{-4} \}$ \\
        Transition model hidden layer size & $\{ 128, 256, 512, 1024 \}$ \\
        Reward model hidden layer size & $\{ 128, 256, 512, 1024 \}$ \\
        CVAE hidden layer size &  $ \{ 256,512,720,1024 \} $ \\
        \bottomrule
    \end{tabular}
    \label{tab:hyperparams:locomotion_dynamics_cvae_range}
\end{table}

% \begin{table}[h]
%     \centering
%     \begin{tabular}{c|c}
%         Hyperparameters & Search Range  \\
%         \hline
%          lr &   [1e-3,1e-4,1e-5,2e-4,53-4] \\
%          hidden neurons in transition model &   [128,256,512,1024] \\
%          hidden neurons in reward model &   [128,256,512,1024] \\
%          \hline
%     \end{tabular}
%     \caption{Grid Search}
%     \label{tab:my_label}
% \end{table}

\subsection{CVAE}
The CVAE consists of an encoder and a decoder, each of which comprises an MLP with two layers of 750 ReLU activated nodes. The latent dimension is same as the action dimension of the environment under consideration. We use Adam optimizer with $l_1$ loss to learn the parameters.

% \begin{table}[h]
%     \centering
%     \begin{tabular}{c|c}
%         Hyperparameters & Search Range  \\
%         \hline
%          lr &   [1e-3,1e-4,1e-5,2e-4,53-4] \\
%          hidden neurons in CVAE &   [256,512,720,1024] \\
%          \hline
%     \end{tabular}
%     \caption{Grid Search}
%     \label{tab:my_label}
% \end{table}
\subsection{Hyper-parameters}

For training the dynamics models, reward model and CVAE, we use a train-validation split of $90:10$. As in the robotics environment, the learning rate and the number of hidden neurons in the networks are decided using grid-search based on the validation loss. The range in which the hyper-parameters are searched is given in Table~\ref{tab:hyperparams:locomotion_dynamics_cvae_range}. 

% The discrepancy threshold and penalty are tuned by training SAC on the CVAE + P-MDP framework and evaluating the resultant agent in the real Gym environment.
%For the locomotion environments, we experiment on two methods of determining the threshold calculation: fraction and quantile.Once the ensemble is trained and fixed, we compute the maximum discrepancy ($m_d$) by taking the maximum discrepancy over all state-action pairs in the offline dataset $\D$. The discrepancy threshold is calculated by either taking a quantile or fraction of the max. discrepancy ($m_d$).

Once the CVAE and the dynamics and reward functions are learned, the discrepancy fraction or quantile of the ensemble, along with the negative reward penalty are tuned by training an SAC policy on the CVAE + P-MDP and evaluating it using the CVAE and the real gym environment, where the actions taken are obtained by passing latent actions through the CVAE decoder. The hyper-parameters thus obtained are given in Tables~\ref{tab:hyperparams:hopper} and \ref{tab:hyperparams:halfcheetah}.

\begin{table}[h]
    \centering
    \caption{CVAE and P-MDP hyper-parameters for Hopper}
    \begin{tabular}{ccc}
    %\hline
    %\multicolumn{3}{c}{\textbf{Hopper Environment } }\\
    %\hline
        \toprule
        & \multicolumn{2}{c}{Dataset} \\
        \cmidrule(r){2-3}
        Parameter & Medium & Medium-Expert \\
        \midrule
        Dynamics model & MLP (512,512) & MLP (512,512) \\
        Activation & ReLU & ReLU \\
        Batch size & 256 & 256  \\
        Learning rate & $10^{-4}$ & $10^{-4}$ \\
        Discrepancy fraction & 1.08 & 0.08 \\
        Negative reward penalty & 20 & 30 \\
        Number of models in ensemble & 5 & 5 \\
        & & \\
        CVAE & MLP(720,720) & MLP(720,720)\\
        Activation & ReLU & ReLU\\
        Batch size & 256 & 128\\
        Learning rate & $10^{-4}$ & $10^{-4}$ \\
        \bottomrule
    \end{tabular}
    \label{tab:hyperparams:hopper}
\end{table}

\begin{table}[h]
    \centering
    \caption{CVAE and P-MDP hyper-parameters for HalfCheetah}
    \begin{tabular}{ccc}
    %\hline
    %\multicolumn{3}{c}{\textbf{HalfCheetah Environment } }\\
        \toprule
        & \multicolumn{2}{c}{Dataset} \\
        \cmidrule(r){2-3}
        Parameter & Medium & Medium-Expert \\
        \midrule
        Dynamics model & MLP (512,512) & MLP (512,512) \\
        Activation & ReLU & ReLU \\
        Batch size & 256 & 256  \\
        Learning rate & $10^{-4}$ & $10^{-4}$ \\
        Discrepancy fraction & 0.1 & 0.0101 \\
        Negative reward penalty & 20 & 20 \\
        Number of models in ensemble & 5 & 5 \\
        & & \\
        CVAE & MLP(720,720) & MLP(720,720)\\
        Activation & ReLU& ReLU \\
        Batch size & 256 & 128 \\
        Learning rate & $10^{-4}$ & $10^{-4}$ \\
        \bottomrule
    \end{tabular}
    \label{tab:hyperparams:halfcheetah}
\end{table}

After fixing the discrepancy threshold and penalty, the hyper-parameters of the MOC and PPO algorithms are set by training on the CVAE + P-MDP model and evaluating using the CVAE and the real gym environment. The hyper-parameters used are given in Table~\ref{tab:hyperparams:locomotion}. 

The learning rate and $\eta$ in MOC are tuned using Ray-Tune. Other hyperparameters are taken from the MOC paper~\cite{Klissarov2021FlexibleOptionLearning}. Similarly, for PPO, all the hyperparameters are tuned except for vf\_coef and ent\_coef and PPO steps which are taken from hugging-face. PPO and SAC have been used from Stable-Baselines3\cite{stable-baselines} and the MOC code has been adopted from its official implementation~\cite{Klissarov2021FlexibleOptionLearning}.

\begin{table}[h]
    \centering
    \caption{Hyper-parameters for MOC and PPO.}
    \begin{tabular}{ccc}
     \toprule
     Hyper-parameter & Hopper-v2 & HalfCheetah-v2 \\
     \midrule
     MOC & ~ & ~ \\
     lr & $10^{-4}, 10^{-4}, 10^{-4}$ & $10^{-4}, 10^{-4}, 10^{-4}$\\
     clipping value & 0.2 & 0.2 \\
     $\eta$ & 0.7 & 0.7 \\
     $ \gamma $  & 0.99 & 0.99\\
     $ \lambda$ & 0.95 & 0.95   \\
     ~ & ~ & ~ \\
     PPO & ~ & ~ \\
     steps & 2048 & 2048\\
     lr & $9.80828 \times 10^{-05}$ & 0.0003\\
     clipping value  & 0.1 &  0.2\\
     vf$\_$coef & 0.835671 & 0.835671 \\
     ent$\_$coef & 0.00229519 & 0.00229519 \\
     \\
     \bottomrule
    \end{tabular}
    \label{tab:hyperparams:locomotion}
\end{table}

\end{document}